\documentclass[11pt]{article} 
\usepackage{acl}

\usepackage{times}
\usepackage{latexsym}
\usepackage[T1]{fontenc}
\usepackage[utf8]{inputenc}

\usepackage{microtype}
\usepackage{inconsolata}
\usepackage{textcomp}
\usepackage{graphicx}
\graphicspath{{figures/}}
\usepackage{tikz}
\usetikzlibrary{positioning,arrows.meta,calc}
\usepackage{hyperref}
\usepackage{url}
\usepackage{booktabs}
\usepackage{multirow}
\usepackage{longtable}
\usepackage{amsmath}
\usepackage{amssymb}
\usepackage{algorithm}
\usepackage{algpseudocode}
\usepackage[most]{tcolorbox}
\usepackage{newfloat}
\usepackage{float}
\usepackage{needspace}
\usepackage{fvextra}

\definecolor{darkblue}{rgb}{0, 0, 0.5}
\hypersetup{
  colorlinks=true,
  citecolor=darkblue,
  linkcolor=darkblue,
  urlcolor=darkblue,
  pdftitle={Can Large Language Models Forecast What Researchers Study Next?},
  pdfauthor={Fenghai Li, Zihan Tang, Haofei Yu, Yining Zhao, Jiaxuan You},
  pdfsubject={Research idea forecasting with large language models},
  pdfkeywords={research forecasting, large language models, scientific discovery}
}

\newcommand{\keyang}[1]{}
\newcommand{\paul}[1]{}

\newcommand{\methodname}{\textsc{MDF}}
\newcommand{\yes}{\textcolor{teal}{\checkmark}}
\newcommand{\no}{\textcolor{red!70}{\ensuremath{\times}}}
\newcommand{\pt}{\textcolor{orange!90!black}{\ensuremath{\sim}}}

\newcommand{\fitbox}[1]{#1}

\colorlet{promptframe}{darkblue}        
\colorlet{promptslot}{darkblue}         
\DeclareFloatingEnvironment[fileext=lop,placement={H},name=Prompt]{prompt}
\fvset{baselinestretch=1.0,samepage=false,fontsize=\footnotesize,%
  breaklines=true,commandchars=\@\|\^}

\newcommand{\slot}[1]{\textcolor{promptslot}{\{#1\}}}

\newcommand{\promptpart}[1]{%
  \par\addvspace{3pt}\nobreak
  {\ttfamily\bfseries\footnotesize\color{promptframe}\MakeUppercase{#1}}%
  \par\nobreak\vspace{-0.4\baselineskip}}

\tcbset{promptstyle/.style={%
  enhanced, colback=white, colframe=promptframe, boxrule=0.7pt, sharp corners,
  left=2.2mm, right=2.2mm, top=1.8mm, bottom=1.6mm,
  fonttitle=\ttfamily\bfseries\footnotesize\color{white},
  coltitle=white, colbacktitle=promptframe, toptitle=0.8mm, bottomtitle=0.8mm,
  fontupper=\ttfamily,
}}
\newtcolorbox{promptboxenv}[1]{promptstyle, title={#1}}
\newtcolorbox{promptboxenvbrk}[1]{promptstyle, breakable,
  title={#1}, title after break={#1\ (continued)},
  toprule at break=0.7pt, bottomrule at break=0.7pt}

\newenvironment{promptbox}[2]{%
  \def\promptboxtitle{#1}\def\promptboxlabel{#2}%
  \par\addvspace{8pt}%
  \begin{promptboxenv}{#1}%
}{%
  \end{promptboxenv}%
  \par\nobreak\vspace{1pt}\captionof{prompt}[\promptboxtitle]{}\label{\promptboxlabel}%
  \par\addvspace{8pt}%
}
\newenvironment{promptboxbrk}[2]{%
  \def\promptboxtitle{#1}\def\promptboxlabel{#2}%
  \par\addvspace{8pt}%
  \begin{promptboxenvbrk}{#1}%
}{%
  \end{promptboxenvbrk}%
  \par\nobreak\vspace{1pt}\captionof{prompt}[\promptboxtitle]{}\label{\promptboxlabel}%
  \par\addvspace{8pt}%
}

\title{Can Large Language Models Forecast What Researchers Study Next?}

\author{
  Fenghai Li\quad
  Zihan Tang\thanks{Work done during a remote internship at UIUC.}\quad
  Haofei Yu\quad
  Yining Zhao\quad
  Jiaxuan You \\[3pt]
  University of Illinois Urbana-Champaign
}

\begin{document}

\maketitle
\begingroup
\renewcommand\thefootnote{}%
\footnotetext{Correspondence to: max7@illinois.edu.}%
\endgroup

\begin{abstract}
Large language models increasingly generate research ideas, yet judging their novelty or feasibility at generation time does not establish whether they anticipate subsequent work. We introduce \textbf{\textsc{IdeaForecastBench}} to evaluate research idea forecasting. Given a community's literature up to a cutoff, a system produces up to five ranked ideas, which are evaluated against later papers. The benchmark comprises $624$ rolling episodes across $52$ topics, with a fixed retrieve-then-judge protocol and separately reported results from two judges. We compare five history-compression strategies across GPT-4.1, Qwen2.5-7B/14B, and Qwen3.5-9B, together with a learned Mode-Decomposition Forecaster (MDF). Under the primary GPT-4.1-mini judge, Summary improves on Direct in Hit@5 and Precision@5 across all four backbones. Qwen2.5 scores above GPT-4.1, whereas Qwen3.5 scores below it. An outcome-blind assessment finds that Qwen2.5 produces broader forecasts, but does not identify how much breadth contributes to its advantage. Threshold and judge diagnostics further clarify the limits of interpreting realization as precise anticipation. \textbf{\textsc{IdeaForecastBench}} provides a common task for studying which research ideas a community subsequently pursues and how reliably this outcome can be measured.
\end{abstract}

\section{Introduction}

\begin{figure}[t]
\centering
\includegraphics[width=\columnwidth]{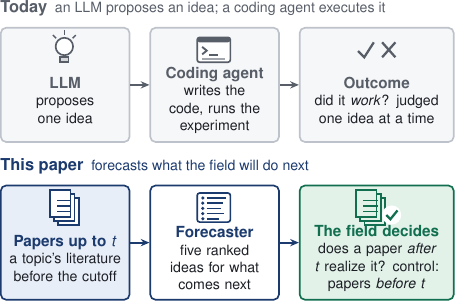}
\caption{\textbf{Executing an idea versus forecasting the field.} Idea execution evaluates a proposal through its own experiment. Idea forecasting asks whether related research is realized in the community's subsequent publications, using only the allowed historical literature as input.}
\vspace{-2mm}
\label{fig:paradigms}
\end{figure}

Large language models (LLMs) are increasingly used for scientific literature understanding \citep{li2024scilitllm}, literature review \citep{tang2025literaturereview}, research ideation \citep{baek-etal-2025-researchagent,si2025novel}, and autonomous discovery \citep{lu2024aiscientist,zheng2025automation}. Research topics and citation impact evolve over time \citep{blei2006dynamic,wang2013impact}, and a community's emerging problems and methods may signal its future work. By compressing this literature, LLMs may capture these signals. We therefore ask: \emph{can LLMs predict research ideas that later emerge in a research community?}

We study this question through \emph{research idea forecasting}. Prior work on research ideation primarily evaluates whether generated ideas are novel, feasible, or promising \citep{si2025novel,baek-etal-2025-researchagent}, or evaluates the outcome of executing selected proposals one at a time \citep{si2025ideation}. We ask whether a forecaster can anticipate the ideas that a research community subsequently pursues (Figure~\ref{fig:paradigms}). Given literature available up to a cutoff $t$, a forecaster produces a ranked list of research ideas, which are evaluated against papers appearing afterward. Subsequent publications provide an observable realization outcome beyond immediate judgments of an idea's persuasiveness.

Research idea forecasting has three potential uses in scientific discovery and the study of research communities. 
\textbf{(1) Early discovery}: a strong forecaster could identify promising directions before they appear in the literature, extending ideation beyond immediate assessments of novelty or plausibility \citep{si2025novel,baek-etal-2025-researchagent,guo2024ideabench}. 
\textbf{(2) Research decision-making}: forecasts could help prioritize problems, methods, and directions for further investigation, informing research planning and autonomous discovery \citep{lu2024aiscientist,zheng2025automation}. 
\textbf{(3) Community modeling}: forecasting provides an operational way to study how research communities evolve, with natural-language ideas as prediction units rather than topics, citations, or aggregate impact \citep{blei2006dynamic,wang2013impact,yu2024researchtown}.

This task presents three challenges. \textbf{(1) Noisy history}: forecasting requires selecting relevant evidence from a large and heterogeneous literature. Under a limited context budget, a forecaster must decide which evidence to retain and how to represent it.
\textbf{(2) Dynamic research directions}: research topics evolve over time \citep{blei2006dynamic}, requiring forecasters to reassess which historical evidence remains informative.
\textbf{(3) Open-ended benchmarking}: research ideas and hypotheses are difficult to evaluate \citep{si2025novel,guo2024ideabench,liu2025hypobench}. Comparing forecasts with future papers grounds evaluation in subsequent research, but there is no single correct future idea, and semantic similarity alone does not establish that a paper realizes a predicted contribution. A broad statement may match many papers while making few testable commitments.


\paragraph{Contributions.}
We make two contributions. \textbf{(1) IdeaForecastBench and its evaluation protocol.} We define community-level idea forecasting and implement it with a shared rolling-window manifest, inspectable idea--paper matching, and diagnostics for sensitivity to the judge and specificity threshold (Section~\ref{sec:problem}). \textbf{(2) Forecasting through history compression.} We organize five prompting strategies by what they retain from history, compare them across four generation backbones, and include the trainable Mode-Decomposition Forecaster (\methodname{}) as a reference implementation (Section~\ref{sec:method}).

\paragraph{Empirical findings.}
Under GPT-4.1-mini judging, Summary improves Hit@5 over Direct from $0.487$ to $0.756$ on GPT-4.1 and from $0.571$ to $0.949$ on Qwen2.5-7B. These realization scores do not establish that nearly every forecast precisely anticipates a new discovery. An outcome-blind assessment of $832$ forecasts finds that Qwen2.5's outputs are broader than GPT-4.1's. Its score advantage persists under a stricter matching gate for several strategies, but that gate is not a control for intrinsic generality. We therefore report the matching advantage without attributing it entirely to either better anticipation or broader wording. The learned reference's scores also vary substantially across judges, motivating diagnostics alongside the leaderboard.

Our aim is to support comparable forecasts and inspectable judgments while tracking changes to benchmark construction, evaluation, and forecasting methods separately.

\section{Related Work}
\label{sec:related}

\paragraph{Automatic research agents.}
Large language models automate code generation from papers \citep{li2025papercoder}, code-based experimentation \citep{jansen2025codescientist}, end-to-end discovery from experiment to paper \citep{lu2024aiscientist}, community-level paper and review generation \citep{yu2024researchtown}, and research idea generation \citep{baek-etal-2025-researchagent,si2025novel,zhao2025ramon}. IdeaBench evaluates generated ideas for novelty and feasibility \citep{guo2024ideabench}; HypoBench evaluates hypotheses for predictive utility, generalizability, and recovery of ground-truth hypotheses \citep{liu2025hypobench}. Neither directly measures whether related ideas later emerge in the literature.

\paragraph{Self-evolving agents.}
Agents that improve across episodes draw on long-term memory \citep{zhong2023memorybank,packer2024memgpt}, reflective feedback \citep{shinn2023reflexion}, embodied skill accumulation \citep{wang2023voyager}, and procedural, reasoning, and evolving-skill memories \citep{cao2025reme,ouyang2025reasoningbank,zhang2026memskill}. These agents are evaluated on conversation, document analysis, question answering, interactive, and coding tasks, whereas our task uses subsequent publications as delayed, inspectable feedback. Our reference forecaster uses a typed memory updated by fixed rules (Appendix~\ref{app:method}).

\paragraph{Forecasting benchmarks.}
Forecasting evaluations cover general future questions \citep{karger2024forecastbench,halawi2024forecasting}, while broader financial benchmarks include prediction and decision-making tasks \citep{xie2023pixiu,xie2024finben}. Closer to science, \citet{wen2025predicting} predict the outcomes of empirical AI experiments without running them; PreScience~\citep{ajith2026prescience} decomposes an advance into structured prediction tasks; and CUSP~\citep{wu2026cusp} forecasts milestone events. We compare the concurrent PreScience and CUSP benchmarks by task design rather than numerical scores. PreScience's contribution-generation subtask is closest to ours: it evaluates a prediction against a single held-out abstract. \textsc{IdeaForecastBench} evaluates a \emph{ranked set} of ideas against a community's post-cutoff paper stream (Table~\ref{tab:positioning}). This shifts the target from reconstructing one abstract or predicting a pre-specified milestone to community-level realization over rolling windows. Because continuations of existing work can also be realized, we distinguish realization from novelty and examine its measurement sensitivity in Section~\ref{sec:judge}.

\begin{table}[t]
\centering
\small
\setlength{\tabcolsep}{3.5pt}
\caption{\textbf{Only \textsc{IdeaForecastBench} scores natural-language ideas against the
full future stream.} All share a time cutoff. \textbf{Idea}: the forecast is a
natural-language idea (\yes/\no/\pt\ = yes/no/partial); \textbf{Scale} counts the questions,
unique in-scope papers, or events a benchmark is built from. IdeaForecastBench's
42.8K is the current deduplicated count assigned to at least one topic.}
\label{tab:positioning}
\fitbox{%
\begin{tabular*}{\columnwidth}{@{\extracolsep{\fill}}l c l r@{}}
\toprule
\textbf{Benchmark} & \textbf{Idea} & \textbf{Target} & \textbf{Scale} \\
\midrule
ForecastBench & \no  & resolved Qs    & 1K \\
PreScience    & \pt  & 1 abstract     & 98K \\
CUSP          & \pt  & milestones     & 4.7K \\
\addlinespace[1pt]
\textbf{\textsc{IdeaForecastBench}} & \yes & \textbf{future stream} & 42.8K \\
\bottomrule
\end{tabular*}}
\end{table}

\begin{figure*}[t]
\centering
\includegraphics[width=\textwidth]{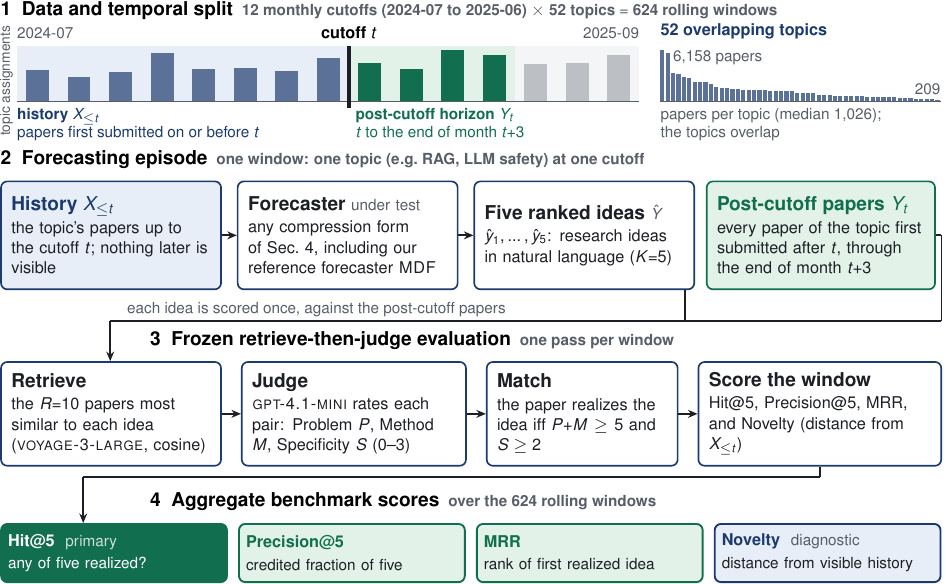}
\caption{\textbf{Overview of \textsc{IdeaForecastBench.}}
(1)~Monthly cutoffs divide historical inputs from post-cutoff target papers across $52$ overlapping topics; the bars show topic assignments over the displayed cutoff intervals, not unique papers.
(2)~Given a topic's history up to $t$, a forecaster produces $K{=}5$ ranked ideas. Targets are papers first submitted after $t$ through the last day of month $t{+}3$.
(3)~For each idea, the evaluator retrieves $R{=}10$ candidate papers and applies the $P{+}M\ge5$, $S\ge2$ matching gate, with at most one credited idea per paper. Judges are reported separately.
(4)~Hit@5, Precision@5, and MRR are averaged over $624$ episodes; historical embedding distance is reported separately as the Novelty diagnostic.}
\label{fig:overview}
\end{figure*}

\section{Benchmarking Idea Forecasting}
\label{sec:problem}
\label{sec:estimand}
\textsc{IdeaForecastBench} evaluates natural-language forecasts against subsequent publications. Its unit is a \emph{topic--cutoff episode}, not an idea with a single pre-specified answer (Figure~\ref{fig:overview}).

\subsection{Problem Definition for Idea Forecasting}
\label{sec:task}
Let $\mathcal{P}_c$ be the papers assigned to topic $c$, $d(p)$ a paper's first-submission date, and $t$ a monthly cutoff date. A forecaster observes the allowed history
\begin{equation}
 X_{c,t}=\{p\in\mathcal{P}_c:d_0\le d(p)\le t\},
\end{equation}
where $d_0$ is the start of the evaluated corpus. It produces an ordered list $\hat Y_{c,t}=(\hat y_1,\ldots,\hat y_K)$ with budget $K=5$. Each idea names a problem and an approach; its structured fields include a title, rationale, method description, and optional key terms. The method may select, summarize, cluster, or otherwise compress the history, but may not access post-cutoff papers as input.

The target is the topic's post-cutoff paper set
\begin{equation}
 Y_{c,t}=\{p\in\mathcal{P}_c:t<d(p)\le e(t)\},
\end{equation}
where $e(t)$ is the last day of the month obtained by adding three to the cutoff month. A forecast is operationally \emph{realized} when a retrieved paper satisfies the idea--paper matching rubric of Section~\ref{sec:protocol}. The task measures whether subsequent work is consistent with the forecast. It does not assess the forecaster's ability to execute the idea, the idea's scientific value, or verbatim agreement with an abstract.

\paragraph{Open-ended targets.}
Many ideas may be realized in one episode, and broad forecasts may resemble several later papers. Because valid future ideas cannot be exhaustively enumerated, we evaluate a fixed forecast budget rather than recall over all possibilities. Interpretation requires both realization frequency and forecast informativeness.

\paragraph{Realization is a proxy.}
Publication provides an inspectable but delayed and incomplete record of realization. Unmatched ideas may appear later or outside the corpus, while matched ideas may be incremental continuations. We measure realization within a specified pool and horizon, not scientific value or uniquely novel anticipation. Section~\ref{sec:experiments} examines the resulting measurement limitations.

\subsection{Data Collection and Benchmark Construction}
\label{sec:construction}\label{sec:data}
\paragraph{Ingestion and deduplication.}
We collect arXiv machine-learning papers with \texttt{cat:cs.ML}, retaining identifiers, titles, abstracts or contribution summaries, submission dates, and category metadata. Repeated results and cross-listings are merged by identifier (Appendix~\ref{app:dataset}).

\paragraph{Temporal assignment.}
We split papers by first-submission date, so later revisions do not move them into the target pool. This rule determines membership; text available at each cutoff requires a separate provenance check. Forecasters select and summarize only historical papers.

\paragraph{Research communities.}
Fixed names, aliases, and keyword rules assign papers to $52$ overlapping topics from title/key-point text, including retrieval-augmented generation, reinforcement learning, and medical imaging. A paper can belong to several topics but appears once within each. Topic definitions remain fixed across methods and cutoffs.

\paragraph{Rolling windows.}
The corpus spans April 2024--September 2025. Twelve monthly cutoffs from July 2024 through June 2025 yield $52\times12=624$ episodes. History expands from April 2024; earlier literature is outside the evaluated snapshot.

\paragraph{Exact calendar boundary.}
For a July 1, 2024 cutoff, history includes that day and targets span July 2--October 31. The three-month endpoint offset thus spans parts of four calendar months. All results share this boundary; overlapping monthly targets remain grouped within topics for uncertainty estimation.

\paragraph{Common episode manifest.}
All $21$ configurations share $624$ topic--cutoff keys, endpoints, and pool sizes. Release verification must also compare paper identifiers, since equal counts do not establish identical membership (Appendix~\ref{app:dataset}).

\paragraph{Availability and scale.}
The eligibility threshold is two historical papers; the actual minimum is $33$. History pools contain $33$--$4{,}530$ papers (mean $589.9$); target pools contain $35$--$1{,}722$ (mean $313.3$). Appendix~\ref{app:dataset} reports all topic-level counts and distinguishes this evaluation slice from the earlier snapshot.

\paragraph{Temporal access versus pretraining.}
Restricting inputs to historical papers does not rule out a pretrained backbone's exposure to target papers. For trained forecasters, training labels and reward papers must also be separated from evaluation targets; earlier training cutoffs alone are insufficient. Appendix~\ref{app:contamination} discusses the available contamination probe and its limitations.

\subsection{Evaluation Protocol for Idea Forecasting}
\label{sec:protocol}
The protocol separates candidate retrieval, rubric-based matching, and aggregation to determine which ideas receive credit. Section~\ref{sec:experiments} reports scores and measurement diagnostics.

\paragraph{Candidate retrieval.}
We embed each forecast and target paper with \textsc{voyage-3-large} ($1024$ dimensions), retrieving the top $R=10$ by cosine similarity. This evaluation-time funnel is distinct from method-side historical retrieval. Similar terminology does not itself establish realization; the judge makes that decision.

\paragraph{Idea--paper matching.}
The judge sees the forecast's title, rationale, approach, and key terms, together with the candidate paper's title and abstract. It scores \emph{Problem} ($P$), \emph{Method} ($M$), and \emph{Specificity} ($S$), each from $0$ to $3$, and provides a short rationale. The default gate is
\begin{equation}
g(\hat y_i,p)=\mathbf{1}[P+M\ge5\ \wedge\ S\ge2].
\end{equation}
The gate requires close problem/method agreement and at least partial realization of the core idea. Raising $S$ from $2$ to $3$ requires closer technical agreement; we examine this sensitivity separately.

\paragraph{Separately reported judges.}
GPT-4.1-mini is primary; Qwen3.5-9B separately evaluates the same predictions. We do not ensemble their scores. Comparisons distinguish execution failures from judgment differences (Section~\ref{sec:judge}).

\begin{figure}[t]
\centering
\includegraphics[width=\columnwidth]{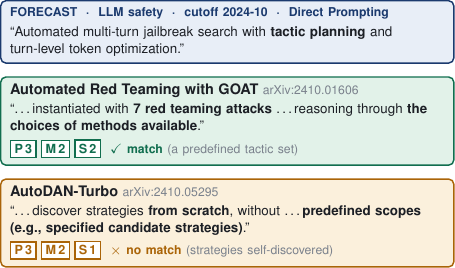}
\caption{\textbf{A worked matching example.} Two papers on the same problem receive different specificity judgments under Qwen3.5-9B. This example comes from the earlier evaluation and illustrates the matching rule; it is not a paired-judge observation from the current experiment.}
\label{fig:episode}
\end{figure}

\paragraph{Crediting a forecast.}
In rank order, each prediction receives credit from the first passing candidate not already credited in that episode. Thus each paper supports at most one idea. Figure~\ref{fig:episode} illustrates rejection; Appendix~\ref{app:eval} details aggregation.

\paragraph{Metrics.}
Let $a_i\in\{0,1\}$ be the credited-match indicator after within-window paper deduplication. We report
\begin{align}
\mathrm{Hit@5}&=\mathbf{1}\!\left[\sum_{i=1}^{5}a_i>0\right],\\
\mathrm{Precision@5}&=\frac{1}{5}\sum_{i=1}^{5}a_i.
\end{align}
Hit@5 asks whether any forecast is realized; Precision@5 measures the fraction of the five-idea budget receiving distinct-paper credit. Missing outputs receive no credit. Mean reciprocal rank (MRR) is $1/\min\{i:a_i=1\}$, or zero if no idea is credited. Table~\ref{tab:main_results} reports all three, averaged over the common episode manifest, alongside judge-independent Novelty. Precision@5 is distinct from the judge's Problem score $P$.

\paragraph{Novelty is a diagnostic.}
For each forecast we also compute its embedding distance from the closest visible historical paper,
\begin{equation}
 N(\hat y_i)=1-\max_{x\in X_{c,t}}\cos(E(\hat y_i),E(x)).
\end{equation}
This judge-independent distance does not measure scientific originality: vague forecasts can be distant from prior work, while useful extensions remain close. We therefore report it separately from realization, without a composite score.

\paragraph{Paired uncertainty.}
We compute $95\%$ percentile intervals from $10{,}000$ topic-clustered bootstrap samples, retaining all cutoffs of each sampled topic. Comparisons pair topic--cutoff episodes. Topic overlap can still induce cross-topic dependence; full intervals appear in Appendix~\ref{app:full-results}.

\paragraph{Changing opportunity sets.}
Larger target pools offer more realization opportunities even with a fixed retrieval depth. We therefore pair methods on identical windows and examine topic consistency and pool-size sensitivity. These checks do not make scores from different snapshots interchangeable.

\section{Forecasting via History Compression}
\label{sec:method}
With the benchmark's history and target pools fixed, we compare how forecasters select and organize evidence. Five \emph{history-compression} strategies use selection, abstraction, trajectories, or memory; MDF introduces a learned structured representation. We compare complete pipelines without equalizing token or compute budgets.

\subsection{Five Forecasting Strategies}
\label{sec:forms}
Table~\ref{tab:strategies} summarizes the five prompting baselines. All condition on the permitted historical side and request the same five-idea output schema. Their prompts and selection rules are retained in Appendix~\ref{app:baselines} and Appendix~\ref{app:prompts}.

\begin{table}[t]
\centering\small
\setlength{\tabcolsep}{3pt}
\caption{\textbf{What each historical representation preserves.} Limits describe the reference implementations; available history can be shorter.}
\label{tab:strategies}
\begin{tabular}{@{}p{.25\columnwidth}p{.69\columnwidth}@{}}
\toprule
\textbf{Strategy} & \textbf{Forecasting context}\\
\midrule
Direct & Up to 20 recent abstracts, without abstraction.\\
Retrieval & Up to 20 historical papers selected for relevance to recent work.\\
Summary & One paragraph distilled from up to 60 recent paper snippets.\\
Topic Trend & A small set of cluster-level research trajectories.\\
Memory & Eight bullets from older work plus up to 20 recent abstracts.\\
\bottomrule
\end{tabular}
\end{table}

\paragraph{Truncation and selection.}
\textsc{Direct} forecasts from recent abstracts in one call, preserving local details but discarding older context. \textsc{Retrieval} instead uses recent titles and keywords to select historical evidence with hybrid semantic and lexical similarity. It changes which papers are retained without abstracting them into a field-level account.

\paragraph{Abstraction and trajectories.}
\textsc{Summary} condenses recent snippets into roughly eight sentences and forecasts from that paragraph alone, testing whether themes and open problems support forecasting without paper-level detail. \textsc{Topic Trend} forecasts from clusters ranked by recent activity. Its output-budget behavior is analyzed in Section~\ref{sec:setup}.

\paragraph{Two-tier memory.}
\textsc{Memory} compresses papers older than six months into eight bullets and combines them with recent abstracts, preserving more short-term detail than Summary. Because the corpus begins in April 2024, early cutoffs lack a substantial older-paper pool. Its effective context therefore changes as history accumulates; the experiment does not provide uniformly long histories.

\subsection{A Learned Reference: MDF}
\label{sec:mdf}
The Mode-Decomposition Forecaster (\methodname{}) introduces a structured intermediate representation of historical innovations. It represents an idea as $z=(b,o,g)$: a base direction $b$, an operator $o$, and a target gap $g$. For example, a base direction could be preference optimization, an operator an extension, and a gap robustness under distribution shift. The operator inventory contains \textsc{extend}, \textsc{transfer}, \textsc{compose}, \textsc{benchmark}, \textsc{analyze}, \textsc{simplify}, \textsc{scale}, and \textsc{adapt}.

\paragraph{From an innovation to an idea.}
A typed memory $\mathcal{M}_t$ stores innovations and their frequency, recency, and utility. The prior $p_\theta(z\mid\mathcal{M}_t)$ predicts an innovation triple, and the realization policy $p_\psi(y\mid z,X_{c,t})$ converts it into a grounded idea. Here, policy \emph{realization} refers to generating idea text; benchmark realization refers to a subsequent paper matching that idea. The decomposition is
\begin{equation}
p(y\mid X_{c,t})\approx\sum_z p_\psi(y\mid z,X_{c,t})\,p_\theta(z\mid\mathcal{M}_t).
\end{equation}
Inference approximates the sum through a candidate pool, blends prior and realization scores, removes near-duplicates, and returns the top five predicted ideas (Algorithm~\ref{alg:inference}). The underlying hypothesis is that predicting structured research moves may be easier than directly generating fully specified ideas.

\paragraph{Learning from delayed realization.}
The prior learns from hindsight-extracted triples; the realization policy uses a gated foresight reward, distinct from evaluation-time retrieval and the P/M/S gate. The evaluated checkpoint uses Qwen2.5-7B. Appendix~\ref{app:method} separates reference training settings from its unverified run manifest. We evaluate MDF as a complete pipeline. Without matched ablations, its score cannot identify the separate contributions of the prior, memory, or reinforcement learning.




\section{Experiments and Analysis}
\label{sec:experiments}
We first compare realization and ranking performance under the shared protocol. We then examine how forecast generality, judge and threshold choices, and MDF's evaluated representation affect the interpretation of these scores.

\begin{table*}[t]
\centering\small
\setlength{\tabcolsep}{3pt}
\caption{\textbf{Forecasting results on 624 common episodes.} Bold marks column maxima, not statistical significance. Both judges use $P+M\ge5$ and $S\ge2$ on the same predictions. Novelty is historical distance, not accuracy. Qwen3.5-9B appears as both a generator and a judge; its generator results use the fixed shard selection in Appendix~\ref{app:qwen35}. Qwen-judge scores remain provisional because of execution failures (Section~\ref{sec:judge}). Intervals and supplementary metrics appear in Appendix~\ref{app:full-results}.}
\label{tab:main_results}
\begin{tabular*}{\textwidth}{@{\extracolsep{\fill}}clccccccc@{}}
\toprule
 & & & \multicolumn{3}{c}{\textbf{GPT-4.1-mini judge}} & \multicolumn{3}{c}{\textbf{Qwen3.5-9B judge}}\\
\cmidrule(lr){4-6}\cmidrule(lr){7-9}
\textbf{Backbone} & \textbf{Strategy} & \textbf{Novelty} & \textbf{Hit@5} & \textbf{Precision@5} & \textbf{MRR} & \textbf{Hit@5} & \textbf{Precision@5} & \textbf{MRR}\\
\midrule
\multirow[c]{5}{*}{GPT-4.1} & Summary & 0.181 & 0.756 & 0.297 & 0.519 & 0.822 & 0.323 & 0.548 \\
 & Memory & 0.160 & 0.729 & 0.267 & 0.456 & 0.772 & 0.292 & 0.464 \\
 & Retrieval & 0.130 & 0.696 & 0.257 & 0.465 & 0.692 & 0.237 & 0.396 \\
 & Topic Trend & 0.238 & 0.625 & 0.135 & 0.598 & 0.655 & 0.148 & 0.626 \\
 & Direct & 0.127 & 0.487 & 0.149 & 0.283 & 0.490 & 0.137 & 0.266 \\
\midrule
\multirow[c]{5}{*}{Qwen2.5-7B} & Summary & 0.189 & 0.949 & 0.528 & 0.686 & 0.968 & 0.563 & 0.707 \\
 & Memory & 0.161 & 0.869 & 0.412 & 0.612 & 0.917 & 0.454 & 0.626 \\
 & Retrieval & 0.112 & 0.769 & 0.338 & 0.524 & 0.837 & 0.372 & 0.562 \\
 & Topic Trend & \textbf{0.297} & 0.745 & 0.153 & 0.742 & 0.779 & 0.163 & 0.773 \\
 & Direct & 0.105 & 0.571 & 0.173 & 0.301 & 0.628 & 0.189 & 0.323 \\
\midrule
\multirow[c]{5}{*}{Qwen2.5-14B} & Summary & 0.201 & \textbf{0.954} & \textbf{0.553} & 0.716 & \textbf{0.973} & \textbf{0.613} & 0.737 \\
 & Memory & 0.178 & 0.913 & 0.440 & 0.610 & 0.962 & 0.504 & 0.648 \\
 & Retrieval & 0.145 & 0.854 & 0.396 & 0.555 & 0.886 & 0.409 & 0.563 \\
 & Topic Trend & 0.288 & 0.784 & 0.167 & \textbf{0.768} & 0.825 & 0.189 & \textbf{0.815} \\
 & Direct & 0.134 & 0.615 & 0.201 & 0.321 & 0.654 & 0.217 & 0.330 \\
\midrule
\multirow[c]{5}{*}{Qwen3.5-9B} & Summary & 0.190 & 0.532 & 0.172 & 0.316 & 0.498 & 0.144 & 0.233 \\
 & Memory & 0.149 & 0.471 & 0.134 & 0.274 & 0.316 & 0.077 & 0.159 \\
 & Retrieval & 0.139 & 0.454 & 0.130 & 0.267 & 0.277 & 0.066 & 0.138 \\
 & Topic Trend & 0.229 & 0.340 & 0.072 & 0.328 & 0.292 & 0.061 & 0.277 \\
 & Direct & 0.138 & 0.226 & 0.057 & 0.133 & 0.131 & 0.028 & 0.062 \\
\midrule
Qwen2.5-7B & MDF & 0.200 & 0.545 & 0.171 & 0.310 & 0.296 & 0.080 & 0.158 \\
\bottomrule
\end{tabular*}
\end{table*}

\subsection{Experimental Setup}
\label{sec:setup}
We evaluate five strategies on GPT-4.1, Qwen2.5-7B/14B, and Qwen3.5-9B, plus MDF: $21$ configurations on $624$ episodes. Each judge scores the same generated predictions. Qwen3.5's overlapping generation shards are resolved by a fixed whole-episode rule (Appendix~\ref{app:qwen35}). Context limits are strategy-specific; token and call budgets are not equalized across strategies.

Table~\ref{tab:main_results} uses final match flags with within-window paper deduplication. Strict-gate analysis uses all candidate judgments; representative P/M/S triples alone cannot reconstruct candidate selection (Appendix~\ref{app:eval}).

\paragraph{Output-budget validity.}
\label{sec:output-validity}
Topic Trend fills five slots in $60.1$--$60.7\%$ of windows on GPT-4.1 and Qwen2.5, but only $12.7\%$ on Qwen3.5. Other strategies fill at least $98.9\%$. All episodes, including empty outputs, remain in the denominator; unfilled slots receive zero credit (Appendix~\ref{app:mdf-diag}).

\subsection{Main Results}
\label{sec:results}\label{sec:levels}

\paragraph{Summary leads in realization frequency.}
Summary has the highest Hit@5 and Precision@5 point estimates for every backbone under both judges. Its primary-judge Hit@5 gains over Direct are $0.269$, $0.378$, $0.338$, and $0.306$ for GPT-4.1, Qwen2.5-7B, Qwen2.5-14B, and Qwen3.5-9B. Memory and Retrieval also improve on Direct. These comparisons measure gains from complete pipelines; they do not isolate abstraction from context selection or additional model calls.

\paragraph{Realization frequency, yield, and rank differ.}
Qwen2.5-14B Summary reaches Hit@5 $0.954$ under the primary judge, but Precision@5 is $0.553$. Thus, almost every episode contains a credited idea, but only about half of the five-slot budget receives distinct-paper credit. MRR is $0.716$, indicating how early the first credited idea appears. The metrics distinguish realization frequency, credited yield, and first-match rank. Precision@5 remains informative when Hit@5 approaches its ceiling; historical embedding distance measures none of these outcomes.

\paragraph{Backbone effects are large but not self-explanatory.}
Qwen2.5 exceeds GPT-4.1 on each strategy, but Qwen3.5 does not share that advantage. Under the primary judge, its Summary Hit@5 is $0.532$, versus $0.756$ for GPT-4.1 and $0.954$ for Qwen2.5-14B; the paired difference from GPT-4.1 is $-0.224$ ($95\%$ CI $[-0.274,-0.173]$). Qwen3.5's Hit@5 is lower on all five strategies. The Qwen2.5 advantage therefore does not extend uniformly across the Qwen family, and these scores do not rank general model capability. The outcome-blind analysis below examines Qwen2.5's breadth but cannot explain Qwen3.5's lower scores.

\paragraph{Compression is not uniformly beneficial in every form.}
Topic Trend exceeds Direct in Hit@5 on all four backbones, but has lower Precision@5 on GPT-4.1 and Qwen2.5. It also exceeds Summary in MRR ($0.598$ vs.\ $0.519$ on GPT-4.1), yet underfills the budget and reuses titles across windows. Its low budget completion on Qwen3.5 further limits attributing score differences to idea quality alone. Interpreting its ranking performance therefore also requires accounting for unfilled slots.

\subsection{Generality and Forecasting Performance}
\label{sec:generality}
Qwen2.5's high scores leave open whether its forecasts anticipate future work more precisely or state broader ideas that more papers can satisfy. We examine forecast breadth and matching opportunities to assess this ambiguity, while keeping their association separate from causal evidence.

\begin{figure*}[t]
\centering
\includegraphics[width=\textwidth]{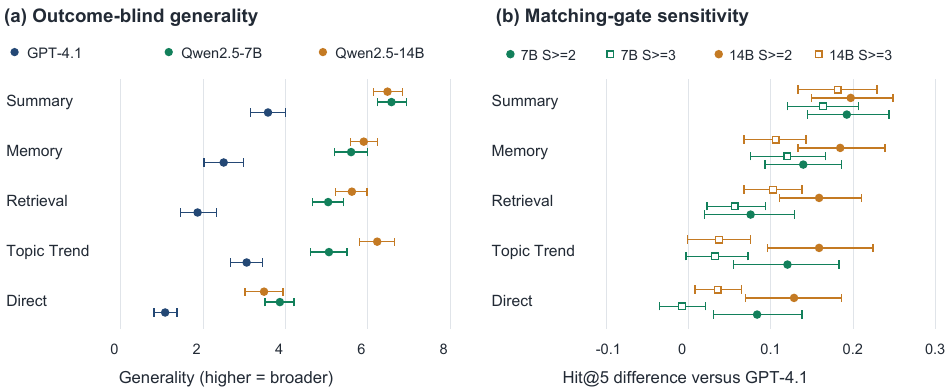}
\caption{\textbf{Broader forecasts and higher realization scores coexist.} (a) Outcome-blind generality on $52$ forecasts per configuration, one per topic. (b) Paired Qwen2.5-minus-GPT-4.1 Hit@5 differences over $624$ windows, using GPT-4.1-mini; 7B and 14B denote Qwen2.5-7B and Qwen2.5-14B. Circles use $S\ge2$; squares use $S\ge3$; bars show $95\%$ topic-clustered intervals. This study covers GPT-4.1 and Qwen2.5, not Qwen3.5. The matching gate does not control intrinsic generality.}
\label{fig:generality}
\end{figure*}

\paragraph{Outcome-blind assessment.}
The blind study samples one forecast per topic from the $15$ GPT-4.1/Qwen2.5 configurations and MDF ($832$ forecasts); Qwen3.5 is not included. GPT-4.1-mini sees only title, rationale, and approach, without future papers, outcomes, backbone labels, or strategy labels. It scores problem, method, and scope specificity, plus testability, from $0$ to $3$. Generality is $12$ minus their sum. The assessment thus measures breadth without candidate papers, although it still relies on an LLM.

\paragraph{Qwen2.5 is consistently broader.}
Figure~\ref{fig:generality}(a) shows higher generality for both Qwen2.5 backbones across all five strategies. On Summary, GPT-4.1 scores $3.58$ ($95\%$ CI $[3.15,4.00]$), compared with $6.58$ ($[6.23,6.94]$) for Qwen2.5-7B and $6.48$ ($[6.13,6.85]$) for Qwen2.5-14B. The pattern supports a systematic difference in how precisely the backbones state their ideas, beyond isolated generic examples.

\paragraph{Breadth and matching opportunities.}
When we link blind ratings to outcomes, forecast-level match rates increase from $0.205$ in the lowest generality bin to $0.406$ in the highest; the correlation is $0.17$ ($0.21$ excluding MDF). Candidate-level evidence is consistent with this association: Summary forecasts have a mean of $0.565$ passing candidates among the retrieved ten for GPT-4.1, versus $1.470$ and $1.642$ for Qwen2.5-7B and 14B. These counts are restricted to the retrieved set and do not cover all future papers an idea could match. The association may also reflect differences in backbone, strategy, and topic (Appendix~\ref{app:generality}).

\paragraph{What remains unresolved.}
Intrinsic specificity concerns how narrowly an idea is stated before outcomes are shown; the matching judge's $S$ concerns how closely a particular paper realizes it. A broad forecast can therefore receive high $S$, and an advantage under $S\ge3$ does not rule out an effect of generality. Limited overlap between the backbones' generality distributions also makes conditional comparisons unstable. We therefore cannot estimate the share of the gap attributable to breadth or interpret a residual as pure forecasting ability. Such attribution requires a controlled specificity intervention or better-matched samples.
\subsection{Judge and Threshold Sensitivity}
\label{sec:judge}
\paragraph{Execution failures versus judge behavior.}
The evaluated artifacts record no judge parse failures for GPT-4.1-mini, but contain failures on the Qwen-judge side. For MDF, all candidate judgments fail in $72$ of $624$ windows. We therefore base the main performance interpretation on GPT-4.1-mini and treat Qwen-judge results as provisional. Until failed judgments are recovered, differences from the primary scores cannot be attributed to judge behavior alone.

\paragraph{Agreement is not accuracy.}
The candidate-level audit reports Summary binary agreement of $0.950$ on GPT-4.1 forecasts and $0.894$ on Qwen2.5-7B forecasts, with Cohen's $\kappa$ of $0.538$ and $0.588$. Most candidate pairs are negatives, so class imbalance partly accounts for the high raw agreement. Agreement alone neither establishes correctness nor rules out shared semantic biases; execution failures further limit interpretation.

\paragraph{Absolute levels change.}
For GPT-4.1 Summary, Hit@5 rises from $0.756$ under GPT-4.1-mini to $0.822$ under the Qwen judge. Qwen3.5 Summary changes in the opposite direction ($0.532$ to $0.498$), as does MDF ($0.545$ to $0.296$). Execution failures limit interpretation, and Qwen3.5's two exports also have minor candidate-set differences (Appendix~\ref{app:qwen35}). These shifts provide no basis for assuming a uniform judging advantage within a model family or applying a universal judge correction.

\paragraph{A stricter gate is a sensitivity analysis.}
We reuse all ten candidates' scores when changing $S\ge2$ to $S\ge3$. On Summary, Qwen2.5-7B's paired Hit@5 advantage over GPT-4.1 changes from $+0.192$ to $+0.163$ ($95\%$ CI $[0.120,0.207]$) under GPT-4.1-mini; the 14B advantage remains $+0.181$ ($[0.133,0.229]$). The advantage persists under the stricter matching rule, but this result does not validate $S$ as a measure of intrinsic specificity or control for forecast breadth. Judges may also differ in how often they assign the highest specificity category.

\paragraph{Human calibration has limited transfer.}
An earlier eight-annotator study includes $340$ pairs and $400$ labels, with core-pair Fleiss' $\kappa=0.135$. It documents disagreement in idea--paper matching on a different experimental slice and therefore does not calibrate the current scores (Appendix~\ref{app:human-agree}). Automated judgments should therefore be interpreted as rubric-based measurements, not human ground truth.

\subsection{MDF Diagnostics}
\label{sec:mdf-results}
MDF achieves Hit@5/Precision@5 of $0.545/0.171$ under GPT-4.1-mini, versus $0.296/0.080$ under Qwen3.5-9B, whose results include execution failures (Section~\ref{sec:judge}). Under the primary judge, MDF trails the stronger prompting strategies and serves as a trainable reference.

\paragraph{Representation loss is not necessarily generation failure.}
The field audit reports a median approach length of four words and empty key-term lists for all $3{,}120$ MDF forecasts. The inspected prediction adapter constructs the approach as \texttt{operator: base\_direction}, leaves key terms at their default, and truncates the rationale to $500$ characters. Conversion can produce sparse evaluation fields even if the raw output contains a method description. Attribution to training requires comparing raw outputs, the deployed adapter, and the final text judged.

\paragraph{What the reference contributes.}
MDF provides a structured, trainable reference and shows why evaluation must preserve information from generated text to scored idea. Component ablations and seed variation remain unmeasured for this checkpoint; older-backbone experiments do not resolve those questions (Appendix~\ref{app:mdf-diag}).

\section{Discussion and Conclusion}
We introduced \textsc{IdeaForecastBench} to study whether LLMs can forecast research ideas realized in a community's subsequent publications. The benchmark compares ranked forecasts on shared topic--cutoff episodes and separately measures realization frequency, credited yield, and first-match rank. Summary achieves the highest Hit@5 and Precision@5 point estimates across the evaluated backbones under both judges. These results show that the representation of historical literature affects how often forecasts align with later papers, but do not by themselves establish precise anticipation of novel scientific contributions.

This distinction remains central to the question posed by our title. Qwen2.5's higher realization scores coexist with broader forecasts, and its advantage under a stricter matching gate does not disentangle breadth from anticipation. Differences across judges and the MDF representation audit further limit a simple capability ranking. We therefore interpret the results as evidence of realization under a specified matching protocol, while leaving precise, novel anticipation unresolved. Better-aligned human calibration, controlled specificity comparisons, and forecasts frozen before target papers appear would strengthen that assessment.

\section*{Limitations}

\paragraph{Automated matching requires further calibration.}
Agreement between GPT-4.1-mini and Qwen3.5-9B does not establish correctness. The auxiliary human study concerns an earlier Qwen-judged experiment and cannot calibrate the current scores (Appendix~\ref{app:blind-recall}). A new study should align human and model instructions, sample current forecasts, and assess missed matches outside the retrieved candidate set.

\paragraph{Judge comparisons remain sensitive to execution and measurement.}
Qwen-judge results include execution failures, including all-candidate failures in $72$ of $624$ MDF windows. Until recovered, score differences cannot be attributed solely to judge behavior (Section~\ref{sec:judge}). Agreement in rankings does not calibrate absolute scores. The outcome-blind generality assessment relies on one LLM, excludes Qwen3.5 generation, and does not identify the causal effect of forecast breadth.

\paragraph{The gate, retrieval depth, and cluster count are operating conventions.}
The protocol retrieves ten candidates and applies $P+M\ge5$, $S\ge2$. Raising the gate to $S\ge3$ tests matching sensitivity without controlling intrinsic specificity. Retrieval bounds the available evidence, while Coverage depends on a five-cluster partition (Appendix~\ref{app:eval}). These choices provide no absolute scale of scientific originality or completeness.

\paragraph{The fixed snapshot does not eliminate pretraining exposure.}
Historical input filtering cannot exclude prior exposure to target papers. The contamination probe is an observational temporal comparison, not a causal estimate of memorization. Historical-text provenance and MDF training-target separation require further audits (Appendices~\ref{app:dataset} and~\ref{app:contamination}). A prospective evaluation would freeze forecasts before collecting their target literature.

\paragraph{The benchmark covers selected machine-learning communities.}
The corpus covers arXiv \texttt{cs.ML} and $52$ overlapping topics. Publications provide an incomplete, delayed record of research; unmatched ideas may be realized later or elsewhere. Cross-domain evaluation must account for publication timing and matching criteria. Topic-clustered intervals retain within-topic dependence, but overlapping membership can induce cross-topic dependence.

\paragraph{\methodname{} is a reference forecaster.}
Token and compute budgets are not equalized across strategies. MDF's sparse evaluated fields may reflect prediction-adapter information loss, rather than generation or training failure (Section~\ref{sec:mdf-results}). Matched adapter comparisons, component ablations, and seed variation remain unavailable. We present MDF as a trainable reference, without attributing its scores to an untested mechanism.

\section*{Acknowledgments}
We gratefully thank Rui Pan and Heng Wang for helpful discussions. We are especially grateful to Paul Liang, whose detailed comments on earlier drafts substantially improved this paper. We also thank the anonymous reviewers for their detailed feedback.

\bibliography{custom}

\clearpage
\onecolumn
\appendix

\section{Dataset Construction and Episode Manifest}
\label{app:dataset}
\subsection{Sources, Dates, and Representations}
The ingestion pipeline queries the arXiv export API with \texttt{cat:cs.ML}, pages through returned entries, and records each paper in a month-bucketed collection. Records retain the canonical identifier, title, first-submission date, category tags, and abstract or contribution summary. Deduplication by arXiv identifier precedes topic assignment. Cross-listed papers are retained once as papers and may acquire more than one topic membership.

The first-submission timestamp is the canonical splitting date. The last-updated timestamp is metadata, not a replacement for the submission date. For cutoff $t$, the historical and future membership tests are mutually exclusive: $d(p)\le t$ and $t<d(p)\le e(t)$, respectively. These membership rules do not establish that the current abstract had the same wording at first submission. A strict prospective release should therefore preserve versioned text and the date of every generated key point, alongside paper membership.

After identifier deduplication, the current \texttt{cs.ML} corpus contains $63{,}855$ papers. Each of the $52$ topics is specified by a name, aliases, and keywords, and assignment is deterministic over the available title/key-point representation. Of the deduplicated papers, $42{,}760$ are assigned to at least one topic and enter benchmark episodes. Topic overlap is intentional; it permits work such as multimodal retrieval to appear in more than one scientifically meaningful context. Consequently, the per-topic counts sum to $71{,}081$ memberships, or $1.66$ memberships per assigned paper. The membership sum is \emph{not} a count of unique papers.

\subsection{Current Evaluation Slice}
The result headers specify \texttt{start\_month=2024-04}, \texttt{end\_month=2025-09}, \texttt{min\_cutoff\_month=2024-07}, \texttt{top\_k=5}, and \texttt{horizon\_months=3}. The last setting is a month-offset parameter: the implemented upper endpoint is the end of cutoff month plus three. The twelve cutoffs are July 2024 through June 2025, each on its first day. The final target therefore ends on September 30, 2025. The configured minimum history is two papers; the actual minimum is $33$.

Across $624$ episodes, historical pools contain $33$--$4{,}530$ papers (mean $589.9$, median $386.5$), and future pools contain $35$--$1{,}722$ (mean $313.3$, median $245.5$). Per-topic corpus sizes range from $209$ to $6{,}158$, with median $1{,}026$. Table~\ref{tab:topic-manifest} supplies the complete topic-level audit. These values are reconstructed from the result manifests, without treating overlapping topic memberships as independent papers.

\paragraph{Relationship to the earlier snapshot.}
An earlier benchmark snapshot covered January 2023--June 2025, with $95{,}276$ ingested papers, $61{,}816$ topic-assigned unique papers, and $1{,}343$ eligible windows. Its main experimental subset contained $208$ windows. Those statistics describe a different corpus and evaluation slice and are not combined with the current main results. The earlier source and figures remain archived to preserve the provenance of the auxiliary studies.

\paragraph{Reproducibility checks.}
For each backbone--strategy--judge result, we verify unique topic--cutoff keys, all $624$ expected episodes, equal historical and future counts across configurations, and agreement between stored match flags and episode-level Hit@5, Precision@5, and reciprocal rank. These checks are necessary but do not establish corpus identity: equal counts cannot replace comparing sorted paper identifiers. Full verification also requires text or content hashes, topic-rule versions, and generation and evaluation configurations.

\small
\begin{longtable}{lrrr}
\caption{\textbf{Current topic manifest.} Every topic has twelve cutoffs. Counts are within-topic paper counts; history and target columns give the minimum--maximum across those cutoffs.}\label{tab:topic-manifest}\\
\toprule Topic & Corpus papers & History range & Target range\\\midrule\endfirsthead
\toprule Topic & Corpus papers & History range & Target range\\\midrule\endhead
3d\_embodied & 2348 & 329--1788 & 468--632 \\
3d\_nerf & 498 & 91--406 & 92--140 \\
active\_learning & 676 & 124--521 & 126--169 \\
anomaly\_detection & 1299 & 191--972 & 267--327 \\
autonomous\_driving & 768 & 136--593 & 146--190 \\
code\_llm & 518 & 72--386 & 95--152 \\
continual\_learning & 1023 & 171--782 & 199--263 \\
dialogue\_conv & 310 & 41--217 & 55--93 \\
domain\_adaptation & 1796 & 316--1415 & 381--421 \\
efficient\_finetuning & 1216 & 180--931 & 244--329 \\
federated\_learning & 2949 & 514--2292 & 608--720 \\
generalization\_ood & 5819 & 794--4295 & 1125--1568 \\
graph\_gnn & 3295 & 569--2556 & 678--786 \\
image\_gen\_diffusion & 1350 & 229--1057 & 265--352 \\
image\_recognition & 1392 & 238--1078 & 289--329 \\
image\_segmentation & 741 & 150--594 & 142--185 \\
in\_context\_learning & 2307 & 375--1727 & 470--603 \\
information\_extraction & 209 & 44--167 & 35--57 \\
knowledge\_distillation & 1366 & 205--1013 & 257--375 \\
knowledge\_graph & 450 & 75--337 & 82--121 \\
llm\_agents & 1409 & 139--940 & 229--469 \\
llm\_alignment\_rlhf & 3418 & 456--2447 & 596--1002 \\
llm\_factuality & 1452 & 211--1047 & 257--405 \\
llm\_instruction & 988 & 131--684 & 178--304 \\
llm\_long\_context & 597 & 73--450 & 103--198 \\
llm\_pretraining & 2085 & 304--1509 & 401--576 \\
llm\_reasoning\_cot & 666 & 57--444 & 90--247 \\
llm\_reasoning\_math & 690 & 65--464 & 107--247 \\
llm\_safety & 478 & 69--369 & 92--138 \\
machine\_translation & 527 & 93--414 & 108--128 \\
medical\_imaging & 1233 & 217--915 & 245--318 \\
medical\_nlp & 264 & 49--189 & 43--78 \\
meta\_learning & 991 & 148--728 & 190--280 \\
moe & 702 & 75--497 & 133--205 \\
molecular\_graph & 724 & 121--550 & 144--179 \\
nas & 297 & 59--232 & 47--82 \\
object\_detection & 433 & 72--329 & 79--110 \\
protein\_structure & 267 & 33--198 & 49--78 \\
pruning\_sparsity & 1342 & 195--1009 & 257--357 \\
quantization & 1211 & 175--911 & 232--339 \\
question\_answering & 765 & 131--596 & 137--210 \\
rag\_retrieval & 733 & 85--564 & 153--221 \\
recommendation & 1029 & 169--769 & 178--264 \\
reinforcement\_learning & 6158 & 978--4530 & 1111--1722 \\
remote\_sensing & 613 & 99--465 & 118--159 \\
scientific\_ml & 1503 & 207--1077 & 285--426 \\
self\_supervised & 2306 & 394--1787 & 481--526 \\
sgd\_training & 1668 & 282--1255 & 328--448 \\
speech\_audio & 585 & 99--425 & 96--160 \\
tabular\_ml & 1567 & 228--1162 & 321--405 \\
time\_series & 2812 & 424--2122 & 587--692 \\
vision\_language & 1238 & 177--945 & 228--323 \\
\bottomrule
\end{longtable}
\normalsize
\section{\methodname{} Architecture and Reference Configuration}
\label{app:method}

\paragraph{Scope of these details.}
This appendix describes MDF's architecture and the earlier Qwen3.5-9B reference configuration. The checkpoint evaluated in the main table instead uses Qwen2.5-7B. The judged artifacts do not include its complete training manifest. The hyperparameters below therefore describe the reference configuration, not verified settings for the evaluated checkpoint. Verification requires checkpoint identifiers, training-target dates, run overrides, and adapter hashes. Earlier ablations are not used to infer component gains for the evaluated checkpoint.

\begin{algorithm}[t]
\caption{\methodname{} joint inference (Section~\ref{sec:method})}
\label{alg:inference}
\begin{algorithmic}[1]
\Require History $X_{c,t}$, memory $\mathcal{M}_t$, pool size $C$, forecast budget $K$, blend $\lambda{=}0.4$, trained prior $p_\theta$, realization policy $p_\psi$
\Ensure Ranked idea list $\hat Y_{c,t}$ with at most $K$ entries
\State $Z \sim p_\theta(z \mid \mathcal{M}_t)$ \Comment{sample $C$ candidate latent innovations from the prior}
\State $S \gets \emptyset$ \Comment{scored ideas}
\For{each $z_i \in Z$}
\State $s_{\text{prior}} \gets |z_i|^{-1}\log p_\theta(z_i \mid \mathcal{M}_t)$ \Comment{mean conditional log-probability per token}
\State retrieve evidence from $X_{c,t}$; \, $\hat y_i \sim p_\psi(y \mid z_i, X_{c,t})$ \Comment{generate an idea from $z_i$}
\State $s_{\text{realize}} \gets |\hat y_i|^{-1}\log p_\psi(\hat y_i \mid z_i, X_{c,t})$ \Comment{mean conditional log-probability per token}
\State $S \gets S \cup \{(\hat y_i,\, \lambda s_{\text{prior}} + (1{-}\lambda) s_{\text{realize}})\}$ \Comment{blend the two scores}
\EndFor
\State \Return $\textsc{DeduplicateAndTopK}(S, K)$ \Comment{drop near-duplicates, keep the top $K$}
\end{algorithmic}
\end{algorithm}

\paragraph{Formulation.}
Let $X_{c,t}$ be the permitted historical papers and $y_j$ an idea-level description of a target paper. MDF introduces the latent innovation $z=(b,o,g)$ of Section~\ref{sec:method} to organize the mapping from history to these descriptions (e.g.\ $b=$ ``preference optimization for alignment'', $o=\textsc{extend}$, $g=$ ``instability of DPO under distribution shift''). A simplifying conditional-independence approximation gives
\begin{equation}
p(y_1,\ldots,y_M \mid X_{c,t}) \approx \prod_{j=1}^{M} \sum_{z_j} p(y_j \mid z_j, X_{c,t})\, p(z_j \mid X_{c,t}),
\end{equation}
which separates a prior over innovations from a policy that expresses each innovation as an idea. In the implementation, the prior conditions on compact memory $\mathcal{M}_t$ rather than the full historical stream. Papers do not provide labels in the required triple schema, so hindsight extraction supplies pseudo-labels $\tilde z$ for supervised training. The prior learns to emit these triples as JSON. At inference, sampling approximates the latent sum, and deduplication produces a ranked list of up to $K$ ideas rather than an exhaustive reconstruction of the target papers. Algorithm~\ref{alg:inference} uses token-normalized scores, with $|z_i|$ and $|\hat y_i|$ denoting scored output-token counts. The training reward uses retrieval depth $R_{\text{rew}}$, distinct from evaluation depth $R$ in Section~\ref{sec:protocol}.
\paragraph{Backbones and adapters.}
The reference configuration uses Qwen3.5-9B for both stages, with LoRA adapters ($r{=}16$, $\alpha{=}32$, dropout $0.05$, target ``all-linear''). The prior uses supervised fine-tuning and the realization policy uses GRPO. A valid held-out temporal evaluation requires training target and reward pools to precede evaluation targets; checking training cutoffs alone is insufficient when horizons overlap. For the evaluated Qwen2.5-7B checkpoint, the training manifest must establish this separation; its score cannot do so.

\paragraph{Hindsight extraction.}
For each paper in a training episode's future pool, a frozen extractor LLM (GPT-5.4, temperature $0.2$, up to $2$ retries) reads its title and abstract, a summary of the selected historical papers, and reference grounding. It returns a triple $\tilde z=(b,o,g)$ as JSON, with the operator checked against the eight allowed values. A second frozen GPT-5.4 call (temperature $0$) assesses whether the triple is entailed by the target abstract and whether the gap can be motivated from the permitted history. Triples failing either check are discarded. This procedure supplies hindsight supervision; it does not imply that training targets were available at the historical cutoff. The extractor user template appears in Prompt~\ref{prompt:hindsight-user}.

\paragraph{Memory.}
$\mathcal{M}_t$ stores typed entries $(b,o,g,\text{frequency},\text{recency},\text{utility})$. New innovations are appended (incrementing frequency on a match); recency decays by $0.9$ per elapsed month; utility is an EMA ($\alpha{=}0.3$) of delayed feedback. The prior conditions on the top-$10$ entries ranked by $0.45\,\text{recency}+0.35\,\widehat{\text{frequency}}+0.20\,\text{utility}$ (frequency normalized by the maximum in the inventory, utility tanh-normalized).

\paragraph{Prior SFT.}
The reference prior is trained for $3$ epochs with learning rate $2\times10^{-5}$, per-device batch size $4$, gradient accumulation $2$, maximum sequence length $4096$, warmup ratio $0.1$, and weight decay $0.01$. The target is the innovation JSON, and the objective is token-level negative log-likelihood.

\paragraph{Realization GRPO.}
For each condition $(z,X_{c,t})$, the policy samples $G$ trajectories $\{y_1,\dots,y_G\}$, scores them with reward $r(y_i)$, and forms group-centered advantages $\hat{A}_i$. In schematic notation, let $\rho_i = p_\psi(y_i\mid z,X_{c,t})/p_{\text{old}}(y_i\mid z,X_{c,t})$ be the importance ratio and $s_i = \min\!\big(\rho_i\hat{A}_i,\,\mathrm{clip}(\rho_i,1{-}\epsilon,1{+}\epsilon)\hat{A}_i\big)$ the clipped surrogate. The objective is
\begin{equation}
\mathcal{J}_{\text{GRPO}}(\psi) = \mathbb{E}\Big[ \tfrac{1}{G}\textstyle\sum_{i=1}^{G} s_i \Big] - \beta\, \mathbb{D}_{\text{KL}}\!\left(p_\psi \,\|\, p_{\text{ref}}\right).
\end{equation}
The reference implementation uses TRL's unmodified \texttt{GRPOTrainer}, with $G{=}8$ generations, KL weight $\beta{=}10^{-3}$, the trainer's default clip range $\epsilon$, learning rate $10^{-5}$, $3$ epochs, per-device batch size $1$, gradient accumulation $2$, maximum prompt/completion lengths $4096/1024$, and LoRA on Qwen3.5-9B ($r{=}16$, $\alpha{=}32$). The setting \texttt{scale\_rewards=\textquotesingle{}none\textquotesingle{}} mean-centers rewards without dividing by the within-group standard deviation, avoiding amplification when reward variance is small. This is a trainer configuration option, not a code modification. The reference runs use seed $0$. A multi-seed comparison is unavailable for the evaluated checkpoint, so episode bootstrap intervals do not quantify training variability.

\paragraph{Reward gates.}
The reference reward $r(y)$ is zero if any of three checks fails. \emph{Well-formedness} requires a valid structured output. \emph{Grounding} requires every cited paper to retrieve a historical neighbor with cosine similarity above $0.3$ in the episode's history index. This check removes citations without a historical neighbor; it does not verify that the neighbor supports the cited claim. \emph{Operator consistency} requires the declared operator, after the collapse map below, to equal the assigned $o$; operators in the \textsc{other} bucket pass without this constraint.

\paragraph{Rubric reward and shaping.}
For a gate-passing rollout, the reward retrieves $R_{\text{rew}}=5$ papers from the training episode's future pool using \textsc{specter} (\texttt{allenai-specter}, $768$ dimensions). This index differs from the $1024$-dimensional Voyage index used in evaluation. A judge scores each rollout--paper pair against the topic rubric, and the maximum score in $[0,1]$ is retained. The judge prompt specifies a one-time $0.2$ penalty for violating a \texttt{must\_not} condition, floored at zero; the code does not subtract it again.

The reward also adds $0.1\times\max(0,\cos\text{-sim})$ against the nearest retrieved future paper. The rubric score is near-binary ($0$ or in $[0.6,1.0]$), so a group with all-zero rubric scores otherwise has no group-centered learning signal. The shaping term therefore supplies a continuous training signal for gate-passing rollouts. It is not used in benchmark evaluation. The rollout prompt appears in Prompt~\ref{prompt:realization}.

\paragraph{Rubric generation and validation.}
A topic rubric $\mathcal{R}$ contains \texttt{criteria} (3--7 matching requirements) and \texttt{must\_not} conditions (0--4 disqualifiers). A frozen GPT-5.4 generates it from positive examples drawn from a training episode's future papers and negative examples drawn from its history. A separate Qwen3.5-9B judge scores the per-topic positive and negative idea--paper pairs, typically a few dozen in each group. Acceptance requires ROC-AUC $\ge0.70$ and no negative scoring at or above the positive median. The implementation calls the latter failures ``leakage hits''; this is a score-separation diagnostic, not evidence that the judge memorized a topic or that training contamination is absent.

An optional rubric-refresh procedure partitions recent rollouts into high- and low-reward pools, regenerates the rubric, and applies the same validation criteria. The procedure replaces the rubric only after validation and records its version history, so changes in training reward can be compared with held-out performance. Refresh is disabled in the reference configuration, which uses a static validated rubric. These training rubrics are distinct from the fixed P/M/S evaluation rubric.

\paragraph{Operators.}
The extractor emits one of the eight inventory operators; for the operator gate and rubric these collapse to a closed set of four (\textsc{limitation-extension}, \textsc{cross-domain-transfer}, \textsc{benchmark-proposal}, \textsc{method-composition}) plus an \textsc{other} bucket. The map is: \textsc{extend} $\to$ \textsc{limitation-extension}; \textsc{transfer}, \textsc{adapt} $\to$ \textsc{cross-domain-transfer}; \textsc{compose}, \textsc{simplify} $\to$ \textsc{method-composition}; \textsc{benchmark} $\to$ \textsc{benchmark-proposal}; and \textsc{analyze}, \textsc{scale} $\to$ \textsc{other}. Operators that fall in \textsc{other} pass the operator gate unconstrained.

\paragraph{Joint inference.}
At inference we sample $C{=}16$ innovations from the prior (temperature $0.8$); for each we retrieve evidence and generate an idea, rank by $0.4\,s_{\text{prior}}+0.6\,s_{\text{realize}}$ (per-token-normalized log-probabilities), deduplicate at Jaccard $0.8$, and return the top $K{=}5$. The full procedure is Algorithm~\ref{alg:inference} above.


\subsection{Prediction Adapter and Information Preservation}
The implementation calls the raw realization text a \emph{proposal}; an adapter converts it into the benchmark's structured idea prediction. The local function \texttt{proposal\_to\_idea\_prediction} takes the first non-thinking line as the title, appends the raw body text to the innovation gap to form a rationale truncated at $500$ characters, and sets the approach to \texttt{operator: base\_direction}. It does not explicitly fill key terms. A short approach field therefore does not establish that the raw output lacks a method. To trace information loss, a release should retain the raw output, parsed fields, and text passed to the embedding model and judge. Attributing such loss to reinforcement learning also requires testing the same adapter on untrained and trained generators.
\section{Forecasting Baselines and Context Budgets}
\label{app:baselines}
The five baselines share a ranked natural-language output schema: title, rationale, approach, confidence, and optional key terms. Confidence is not a calibrated probability and does not replace the explicit output rank. Prompt templates are retained in Appendix~\ref{app:prompts}. Parameters below describe the reference implementation; configuration overrides must be preserved with each run rather than inferred from a strategy name.

\paragraph{Direct Prompting.}
A single forecasting call reads up to $20$ recent historical abstracts. The prompt explicitly states the cutoff and asks for five concrete research ideas. The default forecasting temperature is $0.4$. Raw recency selection is a useful baseline because it retains paper-level detail without spending a separate call on compression.

\paragraph{Summary-Augmented Prompting.}
The first call reads up to $60$ recent paper summaries, each truncated to $300$ characters, and produces a single paragraph of approximately eight sentences. It is asked to capture dominant themes, methodological trajectories, and recurring open problems rather than list individual papers. A second call receives only this paragraph and the cutoff. Default temperatures are $0.3$ for compression and $0.4$ for forecasting. No recent raw abstracts are reintroduced at the second stage, distinguishing this strategy from Memory.

\paragraph{Retrieval-Augmented Prompting.}
The query uses titles and keywords from the recent historical literature. Hybrid semantic and keyword similarity selects up to $20$ papers from the allowed history. The forecasting call then conditions on the selected evidence. This is method-side retrieval from history and must not be confused with the evaluation-time retrieval of post-cutoff candidate papers.

\paragraph{Memory-Augmented Prompting.}
History is split at six months before the cutoff. Up to $60$ older papers, represented by $300$-character snippets, are condensed into eight memory bullets. Forecasting receives those bullets together with up to $20$ recent abstracts. The default compression and forecasting temperatures are $0.3$ and $0.4$. With insufficient older history, the long-term component is correspondingly limited; no additional pre-April-2024 context is assumed in the current slice.

\paragraph{Topic Trend.}
Historical papers are grouped into clusters, which are ranked by recent activity. The generator is asked for ideas associated with the selected clusters. This is a generator-dependent baseline, not a deterministic keyword forecast. The implementation can produce fewer than five usable predictions. We report its observed budget completion without filling missing slots after seeing outcomes.

\paragraph{Fairness and interpretation.}
The compared pipelines vary both information selection and the number of model calls. Their scores therefore do not isolate the effect of compression at equal compute. A controlled comparison would hold input tokens, output tokens, and generation calls fixed while varying only the historical representation. The current benchmark instead evaluates the complete reference strategies under common windows and a common output budget.
\section{Evaluation Protocol and Measurement Details}
\label{app:eval}\label{app:judge-rel}
\subsection{Retrieval, Rubric, and Crediting}
Each prediction is embedded with the same model used for target-paper embeddings (\textsc{voyage-3-large}, $1024$ dimensions). The top $10$ candidates by cosine similarity are evaluated. The prompt includes predicted title, rationale, approach, and key terms and the candidate title and abstract. Temperature is fixed at zero in the reference judge implementation. GPT-4.1-mini and Qwen3.5-9B are reported separately; the latter is identified by the serving alias \texttt{qwen35-9b-judge} in the exported header.

The rubric distinguishes problem agreement, method agreement, and the extent to which the paper realizes the particular proposed idea. Each dimension takes integer values $0$--$3$. A topical similarity without method agreement should not pass $P+M\ge5$. A core idea with differing implementation details can pass $S\ge2$; $S=3$ asks for a closer realization. The archived prompt listings provide concrete anchors, but prompt hashes and server versions should accompany the final release because a nominal judge model does not uniquely identify its evaluation behavior.

Candidates are judged independently. For credit assignment, ideas are processed in rank order, selecting the first passing candidate not previously credited in the same episode. This produces one binary credit per prediction and at most one credit per paper. Hit@5 is one if any credit exists, Precision@5 divides credited predictions by five, and the reciprocal rank is the inverse rank of the first credit, or zero for no credit. MRR averages this value over the complete $624$-episode manifest.

\paragraph{Representative scores are not the candidate set.}
The compact results retain a P/M/S triple from the credited candidate, or from a representative candidate when none is credited. Applying the gate to that triple alone does not reproduce candidate selection. Even a passing candidate may receive no final credit if its paper has already been used. We therefore reconstruct the main table from final match flags and verify it against the stored metrics. Changing gates, retrieval depth, or paper pools requires all candidate judgments or a new retrieval/judging pass, not only the compact per-prediction triples.

\paragraph{Failure handling.}
The compact exports show no judge parse failures for GPT-4.1-mini, but do record failures for Qwen3.5-9B (Section~\ref{sec:judge}); the earlier candidate audit does not establish failure-free execution. Missing outputs receive zero credit under the fixed five-slot budget. This is distinct from a judge failure or a retrieval miss. A reproducible run should log missing outputs, judge failures, and retrieval misses separately, including retries and failures remaining after retries. Affected windows must remain in the denominator. The reference parser and its prompt must be versioned together.

\subsection{Supplementary Metrics}
MRR averages the reciprocal rank of the first credited idea, with zero for episodes without credit. Historical novelty first averages nearest-neighbor embedding distance over emitted ideas within an episode, then averages those episode values. The evaluator assigns an empty episode zero novelty by convention; this is not a judgment about an absent idea's originality. Neither metric replaces joint reporting of Hit@5 and Precision@5.

The reference implementation also records a \textsc{Soft} score: the mean of $(P+M+S)/9$ over credited predictions, set to zero when there are none. It is conditional on credited matches and thus should not be interpreted as independent evidence of forecasting quality. \textsc{Coverage} partitions target-paper embeddings into five KMeans clusters and counts the fraction containing a credited paper. It depends on the chosen clustering and judge. Table~\ref{tab:supplementary-current} reports both metrics and topic-clustered intervals for every configuration and judge; these supplementary diagnostics are not interchangeable with realization or ranking performance.

\subsection{Uncertainty and Sensitivity}
Table reconstruction uses NumPy's seed-$0$ generator to draw $10{,}000$ samples of the $52$ topics with replacement. All twelve cutoffs for a sampled topic remain together. The $2.5$th and $97.5$th percentiles define the interval. With equal episode counts per topic, the mean of topic means equals the overall episode mean. Paired gate-sensitivity analyses also use $10{,}000$ topic resamples and seed $0$, but a different random-number implementation; their last-digit interval endpoints may therefore differ from independently recomputed intervals.

The strict-gate Hit@5 comparisons in Figure~\ref{fig:generality} use all candidates from the server export. No strict-gate precision from its aggregate table is substituted for deduplicated Precision@5: counting every prediction with any passing candidate ignores the one-paper-one-credit rule. Candidate multiplicity is explicitly a different diagnostic and is never labeled precision.

\paragraph{Recall and calibration.}
Retrieval recall cannot be measured by judging only retrieved candidates. Likewise, judge--judge agreement does not measure agreement with human labels or establish a direction of bias. An end-to-end recall study would need to sample beyond retrieved candidates, align human and judge instructions, and specify an estimator for the sampling design. The auxiliary historical human study in Appendix~\ref{app:human-agree} motivates this requirement but is not a new calibration experiment.
\section{Full Current Results}
\label{app:full-results}\label{tab:main_full}\label{app:backbone}
Table~\ref{tab:full-current} reports $42$ judge-specific results: $21$ configurations on $624$ episodes. GPT and Qwen denote GPT-4.1-mini and Qwen3.5-9B judging, distinct from the Backbone column. Qwen-judge values remain provisional (Section~\ref{sec:judge}). Topic-clustered intervals quantify episode-sampling uncertainty; they do not capture variation from training seeds, generator reruns, or shard selection.

Novelty does not use judge decisions. Its independently exported row means differ by less than
$10^{-5}$ and agree at the displayed precision; the main table uses the primary export
for its single Novelty column.

The table excludes earlier $208$-window results, Qwen3.5-9B ablations, and GPT-5.4 generations. These remain archived, not evidence for the current Qwen2.5-7B MDF. A judge shift measured on one archived generator is not applied as a universal correction.

\small
\begin{longtable}{lllcccc}
\caption{\textbf{Default-gate scores and 95\% intervals.} Novelty is historical embedding distance, not a human originality score. Topic Trend retains its incomplete five-slot outputs.}\label{tab:full-current}\\
\toprule Backbone & Strategy & Judge & Hit@5 [CI] & Precision@5 [CI] & MRR & Novelty\\\midrule\endfirsthead
\toprule Backbone & Strategy & Judge & Hit@5 [CI] & Precision@5 [CI] & MRR & Novelty\\\midrule\endhead
GPT-4.1 & Summary & GPT & 0.756 [0.696, 0.814] & 0.297 [0.258, 0.338] & 0.519 & 0.181 \\
GPT-4.1 & Summary & Qwen & 0.822 [0.772, 0.865] & 0.323 [0.288, 0.358] & 0.548 & 0.181 \\
GPT-4.1 & Memory & GPT & 0.729 [0.671, 0.785] & 0.267 [0.230, 0.306] & 0.456 & 0.160 \\
GPT-4.1 & Memory & Qwen & 0.772 [0.729, 0.816] & 0.292 [0.263, 0.323] & 0.464 & 0.160 \\
GPT-4.1 & Retrieval & GPT & 0.696 [0.633, 0.756] & 0.257 [0.219, 0.298] & 0.465 & 0.130 \\
GPT-4.1 & Retrieval & Qwen & 0.692 [0.641, 0.742] & 0.237 [0.209, 0.269] & 0.396 & 0.130 \\
GPT-4.1 & Topic Trend & GPT & 0.625 [0.551, 0.697] & 0.135 [0.118, 0.154] & 0.598 & 0.238 \\
GPT-4.1 & Topic Trend & Qwen & 0.655 [0.596, 0.712] & 0.148 [0.133, 0.163] & 0.626 & 0.238 \\
GPT-4.1 & Direct & GPT & 0.487 [0.420, 0.554] & 0.149 [0.123, 0.176] & 0.283 & 0.127 \\
GPT-4.1 & Direct & Qwen & 0.490 [0.438, 0.543] & 0.137 [0.120, 0.154] & 0.266 & 0.127 \\
Qwen2.5-7B & Summary & GPT & 0.949 [0.921, 0.973] & 0.528 [0.486, 0.569] & 0.686 & 0.189 \\
Qwen2.5-7B & Summary & Qwen & 0.968 [0.949, 0.984] & 0.563 [0.526, 0.599] & 0.707 & 0.189 \\
Qwen2.5-7B & Memory & GPT & 0.869 [0.824, 0.907] & 0.412 [0.371, 0.453] & 0.612 & 0.161 \\
Qwen2.5-7B & Memory & Qwen & 0.917 [0.881, 0.947] & 0.454 [0.417, 0.491] & 0.626 & 0.161 \\
Qwen2.5-7B & Retrieval & GPT & 0.769 [0.708, 0.825] & 0.338 [0.293, 0.384] & 0.524 & 0.112 \\
Qwen2.5-7B & Retrieval & Qwen & 0.837 [0.792, 0.878] & 0.372 [0.335, 0.410] & 0.562 & 0.112 \\
Qwen2.5-7B & Topic Trend & GPT & 0.745 [0.668, 0.817] & 0.153 [0.137, 0.168] & 0.742 & 0.297 \\
Qwen2.5-7B & Topic Trend & Qwen & 0.779 [0.718, 0.835] & 0.163 [0.149, 0.176] & 0.773 & 0.297 \\
Qwen2.5-7B & Direct & GPT & 0.571 [0.511, 0.628] & 0.173 [0.149, 0.197] & 0.301 & 0.105 \\
Qwen2.5-7B & Direct & Qwen & 0.628 [0.577, 0.678] & 0.189 [0.170, 0.209] & 0.323 & 0.105 \\
Qwen2.5-14B & Summary & GPT & 0.954 [0.925, 0.978] & 0.553 [0.511, 0.594] & 0.716 & 0.201 \\
Qwen2.5-14B & Summary & Qwen & 0.973 [0.954, 0.987] & 0.613 [0.575, 0.650] & 0.737 & 0.201 \\
Qwen2.5-14B & Memory & GPT & 0.913 [0.877, 0.944] & 0.440 [0.400, 0.481] & 0.610 & 0.178 \\
Qwen2.5-14B & Memory & Qwen & 0.962 [0.941, 0.979] & 0.504 [0.467, 0.542] & 0.648 & 0.178 \\
Qwen2.5-14B & Retrieval & GPT & 0.854 [0.806, 0.897] & 0.396 [0.348, 0.445] & 0.555 & 0.145 \\
Qwen2.5-14B & Retrieval & Qwen & 0.886 [0.853, 0.918] & 0.409 [0.371, 0.448] & 0.563 & 0.145 \\
Qwen2.5-14B & Topic Trend & GPT & 0.784 [0.720, 0.845] & 0.167 [0.153, 0.182] & 0.768 & 0.288 \\
Qwen2.5-14B & Topic Trend & Qwen & 0.825 [0.774, 0.872] & 0.189 [0.174, 0.205] & 0.815 & 0.288 \\
Qwen2.5-14B & Direct & GPT & 0.615 [0.556, 0.675] & 0.201 [0.172, 0.231] & 0.321 & 0.134 \\
Qwen2.5-14B & Direct & Qwen & 0.654 [0.604, 0.704] & 0.217 [0.193, 0.242] & 0.330 & 0.134 \\
Qwen3.5-9B & Summary & GPT & 0.532 [0.468, 0.598] & 0.172 [0.143, 0.203] & 0.316 & 0.190 \\
Qwen3.5-9B & Summary & Qwen & 0.498 [0.441, 0.558] & 0.144 [0.122, 0.167] & 0.233 & 0.190 \\
Qwen3.5-9B & Memory & GPT & 0.471 [0.407, 0.537] & 0.134 [0.111, 0.158] & 0.274 & 0.149 \\
Qwen3.5-9B & Memory & Qwen & 0.316 [0.264, 0.370] & 0.077 [0.063, 0.091] & 0.159 & 0.149 \\
Qwen3.5-9B & Retrieval & GPT & 0.454 [0.378, 0.530] & 0.130 [0.103, 0.158] & 0.267 & 0.139 \\
Qwen3.5-9B & Retrieval & Qwen & 0.277 [0.228, 0.327] & 0.066 [0.054, 0.078] & 0.138 & 0.139 \\
Qwen3.5-9B & Topic Trend & GPT & 0.340 [0.276, 0.407] & 0.072 [0.058, 0.086] & 0.328 & 0.229 \\
Qwen3.5-9B & Topic Trend & Qwen & 0.292 [0.245, 0.340] & 0.061 [0.050, 0.072] & 0.277 & 0.229 \\
Qwen3.5-9B & Direct & GPT & 0.226 [0.175, 0.282] & 0.057 [0.043, 0.073] & 0.133 & 0.138 \\
Qwen3.5-9B & Direct & Qwen & 0.131 [0.101, 0.165] & 0.028 [0.021, 0.035] & 0.062 & 0.138 \\
Qwen2.5-7B & MDF & GPT & 0.545 [0.470, 0.620] & 0.171 [0.138, 0.206] & 0.310 & 0.200 \\
Qwen2.5-7B & MDF & Qwen & 0.296 [0.234, 0.362] & 0.080 [0.060, 0.102] & 0.158 & 0.200 \\
\bottomrule
\end{longtable}
\normalsize

\subsection{Qwen3.5 Backbone: Selection and Verification}
\label{app:qwen35}
The Qwen3.5-9B export contains generation outputs and separate GPT-4.1-mini and Qwen3.5-9B judgments for all five strategies. Its generation service uses the alias \texttt{gpt-4.1-qwen35}; this is a serving identifier, not the GPT-4.1 backbone. All $624$ topic--cutoff keys, cutoff dates, target endpoints, and historical/target counts agree with the original cohort.

\paragraph{Whole-episode selection.}
Summary contains $768$ episode records with $624$ unique keys. The $144$ overlapping keys have different generated ideas, not simply repeated judgments of the same text. We sort generation filenames lexicographically and retain the first whole episode record, including an empty output when present. This gives precedence to \texttt{fix0--fix2}, then \texttt{m0--m5}, then \texttt{s0--s1}. Both judges use the corresponding selected source shard. Ideas, ranks, and candidate sets are never combined across generation versions. This rule makes the analysis reproducible, but does not establish which run the server intended as canonical. The analysis manifest records per-episode filenames and source-file hashes.

\paragraph{Selection sensitivity.}
Using the last rather than first file for overlapping episodes changes Summary Hit@5 from $0.532$ to $0.543$ under GPT-4.1-mini and from $0.498$ to $0.508$ under Qwen3.5-9B. These values reflect alternative selections of generated outputs, not confidence bounds. Both choices leave Summary below GPT-4.1 and Qwen2.5. We report the first-file convention consistently and retain all $624$ episodes, unlike aggregate summaries that omit empty outputs. The other four strategies have no duplicate episode records.

\paragraph{Execution and candidate alignment.}
The selected outputs contain $141{,}850$ candidate judgments per judge. GPT-4.1-mini has no recorded parse failures; the Qwen judge has $20$ null judgments across $18$ strategy--episode records, with no entirely failed episode (Table~\ref{tab:qwen35-audit}). These nulls receive no credit and remain unresolved. The two judges score matching ranked prediction titles, but $57$ of $14{,}185$ predictions have one candidate replaced between exports. Pair-level agreement must therefore be computed on shared candidate identifiers; the result columns do not compare judges on exactly identical candidate sets. This export does not repair the Qwen-judge failures in the original configurations.

\paragraph{Threshold check.}
Reapplying the stricter $S\ge3$ gate to all stored candidates reduces Qwen3.5 Summary Hit@5 from $0.532$ to $0.050$ under GPT-4.1-mini, and from $0.498$ to $0.191$ under the Qwen judge. The relative ordering of judges therefore reverses with the specificity threshold in this example. These are descriptive, fixed-output sensitivity results, not a calibration of either judge; the Qwen-side null judgments remain unresolved.

\begin{table}[ht]
\centering\small
\caption{\textbf{Qwen3.5-9B generation and evaluation audit.} Every strategy retains $624$ episodes. Failed judgments count candidate pairs, not missing ideas.}
\label{tab:qwen35-audit}
\begin{tabular}{lrrrrr}
\toprule
Strategy & Ideas & Empty episodes & Full five-slot episodes & GPT failures & Qwen failures\\
\midrule
Summary & 3,110 & 2 & 622 & 0 & 7 \\
Memory & 3,095 & 5 & 619 & 0 & 7 \\
Retrieval & 3,085 & 7 & 617 & 0 & 3 \\
Topic Trend & 1,787 & 12 & 79 & 0 & 2 \\
Direct & 3,108 & 0 & 621 & 0 & 1 \\
\bottomrule
\end{tabular}
\end{table}

\clearpage
\subsection{Supplementary Soft and Coverage Scores}
\label{app:supplementary-scores}
Table~\ref{tab:supplementary-current} retains the two additional evaluator outputs for the
same $624$ episodes. Soft averages normalized rubric scores over credited matches within each
window (zero for no match); Coverage measures the fraction of target-paper clusters reached
by credited matches. The former conditions on passing the gate, while the latter depends on
the five-cluster partition. They provide diagnostics, not independent measures of scientific quality.

\begingroup
\footnotesize
\setlength{\tabcolsep}{4pt}
\begin{longtable}{@{}llcccc@{}}
\caption{\textbf{Supplementary scores and 95\% topic-clustered intervals.} Both judges evaluate
the same generated predictions. Topic Trend uses the as-run outputs, without padding.}
\label{tab:supplementary-current}\\
\toprule
 & & \multicolumn{2}{c}{\textbf{GPT-4.1-mini judge}} & \multicolumn{2}{c}{\textbf{Qwen3.5-9B judge}}\\
\cmidrule(lr){3-4}\cmidrule(lr){5-6}
Backbone & Strategy & Soft [CI] & Coverage [CI] & Soft [CI] & Coverage [CI]\\
\midrule\endfirsthead
\toprule
 & & \multicolumn{2}{c}{\textbf{GPT-4.1-mini judge}} & \multicolumn{2}{c}{\textbf{Qwen3.5-9B judge}}\\
\cmidrule(lr){3-4}\cmidrule(lr){5-6}
Backbone & Strategy & Soft [CI] & Coverage [CI] & Soft [CI] & Coverage [CI]\\
\midrule\endhead
GPT-4.1 & Summary & 0.605 [0.554, 0.652] & 0.247 [0.218, 0.277] & 0.700 [0.657, 0.738] & 0.269 [0.243, 0.293] \\
GPT-4.1 & Memory & 0.581 [0.536, 0.627] & 0.221 [0.194, 0.248] & 0.650 [0.612, 0.688] & 0.246 [0.223, 0.271] \\
GPT-4.1 & Retrieval & 0.552 [0.501, 0.601] & 0.202 [0.179, 0.226] & 0.584 [0.541, 0.625] & 0.192 [0.174, 0.211] \\
GPT-4.1 & Topic Trend & 0.497 [0.439, 0.555] & 0.132 [0.116, 0.149] & 0.557 [0.506, 0.607] & 0.144 [0.130, 0.158] \\
GPT-4.1 & Direct & 0.388 [0.335, 0.443] & 0.127 [0.107, 0.149] & 0.417 [0.373, 0.462] & 0.124 [0.108, 0.139] \\
\midrule
Qwen2.5-7B & Summary & 0.765 [0.741, 0.786] & 0.394 [0.368, 0.421] & 0.838 [0.820, 0.853] & 0.424 [0.398, 0.449] \\
Qwen2.5-7B & Memory & 0.698 [0.662, 0.730] & 0.324 [0.295, 0.351] & 0.786 [0.754, 0.814] & 0.355 [0.328, 0.381] \\
Qwen2.5-7B & Retrieval & 0.617 [0.568, 0.661] & 0.251 [0.222, 0.279] & 0.720 [0.679, 0.759] & 0.279 [0.256, 0.303] \\
Qwen2.5-7B & Topic Trend & 0.599 [0.536, 0.659] & 0.152 [0.137, 0.167] & 0.673 [0.618, 0.726] & 0.162 [0.148, 0.174] \\
Qwen2.5-7B & Direct & 0.456 [0.408, 0.503] & 0.149 [0.131, 0.168] & 0.548 [0.503, 0.592] & 0.166 [0.150, 0.182] \\
\midrule
Qwen2.5-14B & Summary & 0.768 [0.744, 0.787] & 0.415 [0.389, 0.440] & 0.838 [0.821, 0.853] & 0.454 [0.429, 0.478] \\
Qwen2.5-14B & Memory & 0.735 [0.706, 0.760] & 0.344 [0.317, 0.372] & 0.827 [0.806, 0.845] & 0.385 [0.361, 0.408] \\
Qwen2.5-14B & Retrieval & 0.685 [0.646, 0.720] & 0.294 [0.266, 0.324] & 0.762 [0.732, 0.789] & 0.313 [0.290, 0.337] \\
Qwen2.5-14B & Topic Trend & 0.633 [0.581, 0.685] & 0.164 [0.150, 0.178] & 0.712 [0.666, 0.757] & 0.184 [0.171, 0.196] \\
Qwen2.5-14B & Direct & 0.493 [0.445, 0.541] & 0.173 [0.151, 0.195] & 0.566 [0.523, 0.609] & 0.192 [0.172, 0.213] \\
\midrule
Qwen3.5-9B & Summary & 0.421 [0.370, 0.473] & 0.153 [0.130, 0.178] & 0.427 [0.377, 0.478] & 0.130 [0.112, 0.148] \\
Qwen3.5-9B & Memory & 0.373 [0.321, 0.425] & 0.121 [0.102, 0.141] & 0.268 [0.224, 0.315] & 0.072 [0.060, 0.086] \\
Qwen3.5-9B & Retrieval & 0.358 [0.298, 0.418] & 0.115 [0.093, 0.138] & 0.238 [0.196, 0.281] & 0.063 [0.053, 0.075] \\
Qwen3.5-9B & Topic Trend & 0.269 [0.218, 0.323] & 0.071 [0.058, 0.085] & 0.249 [0.209, 0.291] & 0.061 [0.050, 0.071] \\
Qwen3.5-9B & Direct & 0.179 [0.138, 0.223] & 0.054 [0.041, 0.068] & 0.113 [0.087, 0.141] & 0.028 [0.021, 0.035] \\
\midrule
Qwen2.5-7B & MDF & 0.427 [0.367, 0.485] & 0.138 [0.115, 0.161] & 0.253 [0.199, 0.309] & 0.071 [0.054, 0.089] \\
\bottomrule
\end{longtable}
\endgroup
\section{Outcome-Blind Generality Analysis}
\label{app:generality}
\subsection{Sampling and Rubric}
The outcome-blind study covers GPT-4.1 and the two Qwen2.5 backbones across five strategies, plus MDF: $16$ configurations, excluding Qwen3.5 generation. Using seed $0$, it samples one prediction per topic per configuration, giving $52$ predictions per configuration and $832$ overall. Sampling spans cutoffs and ranks rather than selecting only successful predictions. GPT-4.1-mini at temperature zero receives only title, rationale, and approach; future papers, match decisions, model identity, and strategy identity are withheld. Withholding these fields controls the information presented, but writing style may still reveal model identity. The addition of Qwen3.5 to the main table does not extend this blind sample.

Four dimensions are scored from $0$ to $3$: (i) problem specificity, from a broad area to a precise task and failure mode; (ii) method specificity, from no concrete mechanism to an architecture, algorithm, or procedure; (iii) scope specificity, from no setting to a stated dataset, domain, modality, or scale; and (iv) testability, from no stated check to a measurable outcome or comparison. Generality is $12$ minus the sum. We retain the full dimension-wise results in Table~\ref{tab:blind-generality} rather than treating word count as a sufficient measure of specificity.

\begin{table}[ht]
\centering\small
\caption{\textbf{Outcome-blind assessment.} The first four scores measure specificity ($0$--$3$ each); the final column is generality ($12$ minus their sum). Each row has one sampled forecast from every topic.}\label{tab:blind-generality}
\begin{tabular}{llrrrrrr}
\toprule Backbone & Strategy & $n$ & Problem & Method & Scope & Testability & Generality\\\midrule
GPT-4.1 & Summary & 52 & 2.52 & 2.00 & 1.79 & 2.12 & 3.58 \\
GPT-4.1 & Memory & 52 & 2.60 & 2.21 & 2.29 & 2.40 & 2.50 \\
GPT-4.1 & Retrieval & 52 & 2.83 & 2.44 & 2.29 & 2.58 & 1.87 \\
GPT-4.1 & Topic Trend & 52 & 2.42 & 2.04 & 2.12 & 2.37 & 3.06 \\
GPT-4.1 & Direct & 52 & 2.94 & 2.42 & 2.58 & 2.98 & 1.08 \\
Qwen2.5-7B & Summary & 52 & 1.87 & 1.52 & 0.98 & 1.06 & 6.58 \\
Qwen2.5-7B & Memory & 52 & 2.00 & 1.75 & 1.27 & 1.38 & 5.60 \\
Qwen2.5-7B & Retrieval & 52 & 2.10 & 1.83 & 1.52 & 1.52 & 5.04 \\
Qwen2.5-7B & Topic Trend & 52 & 2.00 & 1.67 & 1.48 & 1.79 & 5.06 \\
Qwen2.5-7B & Direct & 52 & 2.23 & 1.88 & 1.75 & 2.27 & 3.87 \\
Qwen2.5-14B & Summary & 52 & 1.81 & 1.54 & 1.06 & 1.12 & 6.48 \\
Qwen2.5-14B & Memory & 52 & 2.02 & 1.67 & 1.27 & 1.13 & 5.90 \\
Qwen2.5-14B & Retrieval & 52 & 2.08 & 1.81 & 1.33 & 1.17 & 5.62 \\
Qwen2.5-14B & Topic Trend & 52 & 1.83 & 1.37 & 1.17 & 1.40 & 6.23 \\
Qwen2.5-14B & Direct & 52 & 2.38 & 2.00 & 1.81 & 2.33 & 3.48 \\
MDF & MDF & 52 & 1.50 & 1.06 & 0.90 & 0.56 & 7.98 \\
\bottomrule\end{tabular}
\end{table}

\subsection{Associations and Common Support}
The outcome-linked summary groups $832$ assessed forecasts into generality bins $[0,3)$, $[3,5)$, $[5,7)$, and $[7,13)$, with sample sizes $190$, $170$, $329$, and $143$. Their match rates are $0.2053$, $0.2294$, $0.3465$, and $0.4056$. This is a forecast-level association, not Hit@5, and it must not be plotted as four episode-level benchmark scores. The overall correlation is $0.17$, or $0.21$ without MDF. Individual forecast identifiers linking the ratings to outcomes should be retained with the analysis; aggregate ratings alone do not reconstruct that join.

GPT-4.1 and Qwen2.5 have markedly different generality distributions. Within a narrow generality band, the backbones share little overlap and few comparable forecasts. Reweighting Qwen2.5 forecasts to the GPT-4.1 distribution is therefore unstable. We do not report a causal decomposition or a percentage of the backbone gap explained by generality. A stronger test would produce paired variants of the same forecast with differing specificity, independently verify that their core content is preserved, and evaluate them on the same candidates without outcome-informed rewriting.

\subsection{Matching Breadth and Gate Sensitivity}
For each prediction, retrieved match multiplicity is the count of its ten retrieved candidates passing the gate. This is a truncated lower bound on matches in the target pool. Mean Summary multiplicity under the primary judge is $0.565$, $1.470$, and $1.642$ for GPT-4.1, Qwen2.5-7B, and Qwen2.5-14B. These descriptive means include zero-match predictions. The supplied paired-multiplicity implementation skips some zero-match comparisons; its paired differences and intervals are excluded from this paper.

The stricter-gate contrasts in Figure~\ref{fig:generality} instead concern the existence of any passing candidate per window and retain zero-hit windows. These contrasts assess sensitivity to the matching rule, but do not identify the effect of intrinsic specificity. In particular, the matching rubric evaluates realization \emph{relative to a candidate paper}; it does not independently measure how many commitments the forecast made before the paper was shown.

Pool-size stratification is also descriptive. At high Hit@5, a larger pool may produce a ceiling effect, reducing between-model gaps even when both gain more matching opportunities. We therefore do not use a narrowing gap at larger pools to reject the generality explanation. Using shared windows holds the target pool fixed across models, but does not hold forecast breadth fixed.
\section{Output Validity and Learned-Forecaster Diagnostics}
\label{app:mdf-diag}\label{app:ablation}
\subsection{Budget Completion and Reuse}
The original $16$ configurations have no empty prediction windows; the five Qwen3.5 strategies add $2$, $5$, $7$, $12$, and $0$ empty windows for Summary, Memory, Retrieval, Topic Trend, and Direct, respectively. None of the $21$ configurations has within-window identical-title duplicates. Table~\ref{tab:qwen35-audit} reports Qwen3.5's output counts, and Table~\ref{tab:output-validity} compares Topic Trend across backbones. Repetition across adjacent windows is not inherently invalid because history and targets overlap. We report it as a diagnostic, not a reason to remove outputs after observing their scores.

\begin{table}[ht]
\centering\small
\caption{\textbf{Topic Trend output validity.} Reuse counts repeated titles beyond their first occurrence across the run; it is not within-window duplication.}\label{tab:output-validity}
\begin{tabular}{lrrrr}
\toprule Backbone & Predictions & Full-budget windows & Repeated titles & Reuse rate\\\midrule
GPT-4.1 & 2,738 & 379/624 & 831 & 30.4\%\\
Qwen2.5-7B & 2,726 & 375/624 & 1,270 & 46.6\%\\
Qwen2.5-14B & 2,738 & 379/624 & 1,205 & 44.0\%\\
Qwen3.5-9B & 1,787 & 79/624 & 133 & 7.4\%\\
\bottomrule\end{tabular}
\end{table}

The fixed-budget evaluation assigns zero credit to unfilled slots. Precision over emitted ideas would answer a different question and could favor returning a smaller, selectively chosen set. It therefore cannot replace Precision@5 without changing the task. We report budget completion alongside the fixed-budget results.

\subsection{MDF: Observations versus Causes}
MDF emits $3{,}120$ forecasts, with all $624$ windows filling the five-slot budget. The server's field audit reports median approach length $4$ words, $90$th percentile $5$, and empty key-term lists for every forecast. The blind assessment gives generality $7.98$ ($95\%$ CI $[7.50,8.46]$), above the prompting configurations. These measurements describe the evaluated representation; they do not directly assess the raw generated text.

The judge disagreement is substantial: default-gate Hit@5 is $0.545$ under GPT-4.1-mini and $0.296$ under Qwen3.5-9B. Candidate-level specificity-zero rates reported in the server audit are $58.54\%$ and $92.99\%$, respectively. Recorded Qwen-judge failures (Section~\ref{sec:judge}) can contribute zero scores, so this distribution cannot yet isolate semantic judge disagreement. The conversion behavior in Appendix~\ref{app:method} also prevents attributing sparse fields directly to training failure.

Three comparisons would help separate causes without retraining: (i) raw realization output versus final prediction fields, (ii) the deployed adapter versus the inspected implementation, and (iii) judgments of the same generated idea before and after conversion. If raw output is already underspecified, the limitation precedes conversion; if information is lost only in the adapter, evaluation input construction is implicated. The compact judged export does not support these comparisons, so no counterfactual score is claimed here.

\subsection{Status of Component Ablations}
Earlier ablations used a Qwen3.5-9B configuration and do not establish the effect of the prior, reward, or memory in the evaluated Qwen2.5-7B checkpoint. A matched ablation would hold training data, backbone, conversion, decoding, and evaluation fixed while removing one component. Such experiments are unavailable for this checkpoint; the earlier tables remain archived as results of their original configuration.
\section{Auxiliary Human Calibration Study}
\label{app:human-agree}
\paragraph{Scope.}
This auxiliary study documents the difficulty of idea--paper matching and the effects of annotation design. It samples an earlier $208$-window experiment evaluated with Qwen3.5-9B, not the current $624$-window evaluation. Its precision and recall estimates therefore characterize that historical sample and do not calibrate the current main table.

\subsection{Superseded Single-Annotator Study}
The original sampling frame pooled $51{,}770$ candidate pairs from five model--strategy rows, with a judge-match prevalence of approximately $1.7\%$. To avoid an almost-all-negative uniform sample, the study selected $80$ pairs: $40$ judge-matches, $25$ borderline cases, and $15$ clear negatives. An expert first labeled the pairs without the judge's output and then reconsidered disagreements after seeing its judgment and rationale. Showing the automated rationale during reconciliation may shift labels toward the judge's decision.

\begin{table}[ht]
\centering\small
\caption{\textbf{Historical single-annotator study, superseded by the blind study below.} These counts document the original sample, not current judge accuracy.}\label{tab:human-agree}
\begin{tabular}{lrr}\toprule Judge stratum & Pairs & Human match\\\midrule
Match & 40 & 36\\
No-match, borderline & 25 & 3\\
No-match, clear & 15 & 3\\\midrule
Total & 80 & 42\\\bottomrule\end{tabular}
\end{table}

The resulting $36/40=0.90$ agreement on judge-matches is an optimistic reconciled estimate and is not cited as the judge's precision. Stratified sampling requires correct prevalence weights for population estimates. A single annotator also provides no inter-annotator agreement measure.

\needspace{8\baselineskip}
\subsection{Blind Multi-Annotator Study}
\label{app:blind-recall}
The follow-up study used eight annotators, withheld judge outputs, and included negative strata to permit weighted recall estimation. Across $340$ distinct labeled pairs and $400$ annotator-labels, the historical analysis reported precision $0.456$ and pooled recall $0.043$ ($95\%$ CI $[0.026,0.067]$). Recall used a Horvitz--Thompson estimator over sampling strata and a cluster bootstrap over pairs. Thirty core pairs had at least two raters; Fleiss' $\kappa$ was $0.135$ and Krippendorff's $\alpha$ was $0.144$.

\paragraph{Interpretation.}
The large change from the reconciled study illustrates sensitivity to annotation procedure. Low inter-rater agreement also limits confidence in treating one set of labels as definitive. Annotators received brief instructions, whereas the automated judge had anchored rubric levels and examples. Disagreement may therefore reflect different instructions or interpretations of the task, as well as model errors. These results motivate better-matched annotation protocols; they do not imply a universal correction for automated realization scores or that any difference between two scores cancels judge bias.

\paragraph{Participants and materials.}
The eight annotators were authors and research-group members participating voluntarily; no external participants were recruited and no compensation was provided. They labeled prediction text and public paper titles/abstracts. The study collected no personal data about participants. Sampling strata, frozen assignments, blind identifiers, and instructions are retained with the study materials. A new calibration study should sample the current generator outputs and both current judges, preserve the same rubric for people and models, and assess missed matches beyond the retrieval funnel if end-to-end recall is the objective.
\section{Reproducibility, Contamination, and Release Scope}
\label{app:repro-ethics}
\subsection{Artifact Separation}
The main results, all-candidate analyses, and historical studies serve different purposes. Compact judged files contain the final match flags, matched identifiers, ranks, and episode metrics needed to reconstruct the main table. All-candidate judgments support retrieval-set multiplicity and gate sensitivity. Raw model outputs and training manifests are needed to diagnose MDF and audit training-target separation. Distinguishing these artifact types prevents historical results or incomplete exports from being used to support claims they cannot establish about the current evaluation.

Each release should identify the corpus snapshot, canonical paper identifiers and date rule, topic assignments, cutoff manifest, generator checkpoint and prompt, decoding configuration, conversion adapter, embedding model and dimension, judge prompt and serving model, retrieval depth, gate, and metric version. A cached judgment is reusable only when the prediction, candidate text, and judge configuration are unchanged. A completed-window flag is not a substitute for such a cache key.

\subsection{Contamination Probe}
\label{app:contamination}
A within-GPT-4.1 Summary probe compares earlier windows with the first six cutoffs of the current run. It attempts to align historical duration and uses the reported model knowledge boundary to distinguish potentially exposed from later windows. Under the default gate, Hit@5 is $0.7436$ versus $0.7276$, a difference of $+0.0160$ with $95\%$ interval $[-0.053,+0.085]$. This is an observational temporal comparison, not a randomized test of memorization.

The two groups also differ in historical and future paper counts and possibly topic difficulty. With these confounders unresolved, an interval for the observed difference cannot bound the causal effect of contamination. A model's stated knowledge boundary does not enumerate all training documents, and conclusions for GPT-4.1 do not establish that Qwen2.5 is unexposed. We consequently do not use this probe to label the entire current benchmark contamination-free or to explain away the backbone gap.

\paragraph{Training-specific leakage.}
For MDF, hindsight triples and reward pools can contain post-cutoff papers by design during training. This is legitimate supervision only if those target papers do not overlap evaluation targets. The audit must inspect target dates and identifiers, not just the dates at which training episodes begin. The compact evaluation package does not establish that separation by itself.

\subsection{Prospective Evaluation}
A stronger future release would timestamp model weights, prompts, historical inputs, and generated forecasts before the target papers are collected. After the horizon closes, the frozen forecasts would be evaluated under a predeclared protocol. Freezing forecasts in advance would reduce exposure through retrospective generation, while publication lag, corpus coverage, and judge calibration would remain limitations. The present paper reports a retrospective benchmark and does not claim to have performed this prospective experiment.

\subsection{Research Use and Assistance}
The benchmark uses public scholarly metadata and text. Redistribution remains subject to the provenance and licensing of the underlying records; code licensing does not determine rights in third-party abstracts. The human study uses voluntary research-group annotations of scientific text, as described in Appendix~\ref{app:human-agree}.

AI assistants supported code inspection, data consistency checks, figure preparation, and manuscript drafting. Authors remain responsible for the scientific claims, experimental provenance, and released materials. Main-table confidence intervals are computed offline from stored evaluation results, without additional model generation or judging.
\section{Detailed Prompts}
\label{app:prompts}

The reference prompt templates are organized into three groups: \methodname{} training prompts
(Prompts~\ref{prompt:hindsight-sys}--\ref{prompt:realization}; Appendices~\ref{app:dataset}
and~\ref{app:method}), the five baseline forecasting prompts
(Prompts~\ref{prompt:topic}--\ref{prompt:memory}; Appendix~\ref{app:baselines}), and the
retrieve-then-judge evaluation prompts
(Prompts~\ref{prompt:judge-sys}--\ref{prompt:judge-user}; Appendix~\ref{app:eval}). Exact prompt hashes and any server-side overrides for the new runs should accompany the final release; retaining a template is not a verification that every deployed prompt was identical.

In each box, the navy bar names the prompt; navy bold headers mark the role of each
segment (\texttt{\textbf{SYSTEM}}, \texttt{\textbf{USER}}, or a call-stage label such as
\texttt{\textbf{COMPRESS/USER}}); and curly-brace tokens shown in navy, such as
\slot{cutoff\_month}, are runtime placeholders filled per forecasting episode.

\needspace{6\baselineskip}
\subsection{\methodname{} Training Prompts}

\begin{promptbox}{Hindsight innovation-extraction system prompt}{prompt:hindsight-sys}
\promptpart{SYSTEM}
\begin{Verbatim}[breaklines=true]
You are a research analyst specializing in identifying conceptual leaps in scientific
literature. Given a future research paper and its historical context, you extract the
core innovation as a structured triple.
The innovation triple has three components:
- base_direction: The foundational research direction or prior work this paper builds
  upon (2-4 words, e.g. "diffusion language models", "in-context learning")
- operator: The methodological innovation applied (one of: extend, transfer, compose,
  benchmark, analyze, simplify, scale, adapt)
- gap: The specific problem or limitation this paper addresses (1-2 sentences)
Always return ONLY a valid JSON object with exactly these three string fields. No
explanation, no markdown.
\end{Verbatim}
\end{promptbox}

\begin{promptbox}{Hindsight extraction user template}{prompt:hindsight-user}
\promptpart{USER}
\begin{Verbatim}[breaklines=true]
Historical context (papers available before the cutoff):
@slot|context_summary^
Future paper to analyze:
Title: @slot|future_title^
Abstract: @slot|future_abstract^
Reference grounding:
@slot|future_grounding^
Extract the innovation triple for this future paper given the historical context.
Return JSON: {"base_direction": "...", "operator": "...", "gap": "..."}
\end{Verbatim}
\end{promptbox}

\begin{promptbox}{Realization-policy rollout}{prompt:realization}
\promptpart{SYSTEM}
\begin{Verbatim}[breaklines=true]
You are a research scientist skilled at developing detailed research proposals. Given a
conceptual innovation and supporting evidence from prior work, write a comprehensive
research proposal. A good research proposal includes:
- A specific, descriptive title
- The core research question or objective
- The methodology and technical approach
- Why this direction is promising given recent trends
- Expected contributions and outcomes
Think step-by-step about which evidence to cite and which operator best addresses the
gap before writing the proposal. After closing your reasoning, output the proposal as
plain text (title on first line, then the body).
\end{Verbatim}
\promptpart{USER}
\begin{Verbatim}[breaklines=true]
Innovation to realize:
- Base direction: @slot|base_direction^
- Operator: @slot|operator^
- Gap being addressed: @slot|gap^
Historical context available before the cutoff:
@slot|context_summary^
Supporting evidence from prior work:
@slot|evidence_summary^
Write a research proposal for this innovation using only the historical context and
evidence above. Be specific and technically grounded.
Return as plain text (title on first line, then the proposal body).
\end{Verbatim}
\end{promptbox}

\needspace{6\baselineskip}
\subsection{Baseline Forecasting Prompts}

\begin{promptbox}{Topic Trend baseline (one call per top-ranked cluster)}{prompt:topic}
\promptpart{SYSTEM}
\begin{Verbatim}[breaklines=true]
You are a precise research forecasting assistant. Return only valid JSON matching the
requested schema.
\end{Verbatim}
\promptpart{USER}
\begin{Verbatim}[breaklines=true]
You are a research trend forecaster. The following keywords define a growing research
topic cluster (literature cutoff: @slot|cutoff_month^):
  Keywords: @slot|keyword_list^
Based on the trajectory of this topic, generate @slot|n_ideas^ concrete research idea(s)
likely to appear in the months following @slot|cutoff_month^. Do NOT use knowledge from
after @slot|cutoff_month^.
Return JSON only:
[{"title": "...", "rationale": "...", "approach": "...", "confidence": 0.0, "key_terms": []}]
\end{Verbatim}
\end{promptbox}

\begin{promptbox}{Direct Prompting baseline}{prompt:direct}
\promptpart{SYSTEM}
\begin{Verbatim}[breaklines=true]
You are a research forecasting assistant.
Generate concrete, next-stage ideas that can be tested in the near term.
Keep each idea specific, technical, and non-generic.
\end{Verbatim}
\promptpart{USER}
\begin{Verbatim}[breaklines=true]
You are given recent abstracts from the domain: @slot|domain^.
Forecast @slot|n_ideas^ plausible research ideas likely to appear in @slot|horizon^.
The literature cutoff is @slot|cutoff_month^; do not use knowledge from after that cutoff.
Abstracts:
@slot|abstracts^
Return JSON only in the following format:
{
  "ideas": [
    {
      "title": "short title",
      "rationale": "why this follows from the abstracts",
      "approach": "how to implement or test it",
      "score": 0.0,
      "confidence": 0.0,
      "key_terms": ["optional", "keywords"]
    }
  ]
}
`score` and `confidence` must be floats between 0.0 and 1.0.
\end{Verbatim}
\end{promptbox}

\begin{promptbox}{Summary-Augmented Prompting baseline}{prompt:summary}
\promptpart{COMPRESS/SYSTEM}
\begin{Verbatim}[breaklines=true]
You are a research summarizer. Output a single short paragraph; no preamble, no bullets.
\end{Verbatim}
\promptpart{COMPRESS/USER}
\begin{Verbatim}[breaklines=true]
The following papers were published before @slot|cutoff_month^.
@slot|paper_block^
Compress this historical literature into a SHORT single-paragraph summary of about 8
sentences. Capture dominant research themes, methodological trajectories, and recurring
open problems. Do not list individual papers; produce a coherent prose summary.
\end{Verbatim}
\promptpart{FORECAST/SYSTEM}
\begin{Verbatim}[breaklines=true]
You are a research forecasting assistant working from a compressed historical summary.
Return only valid JSON matching the requested schema.
\end{Verbatim}
\promptpart{FORECAST/USER}
\begin{Verbatim}[breaklines=true]
Historical literature summary (cutoff @slot|cutoff_month^):
@slot|summary^
Working ONLY from this summary (do not invoke knowledge from after @slot|cutoff_month^),
forecast @slot|top_k^ concrete research ideas likely to appear in the months following
@slot|cutoff_month^.
Return JSON only:
{"ideas": [{"title": "...", "rationale": "...", "approach": "...", "confidence": 0.0,
            "key_terms": []}]}
\end{Verbatim}
\end{promptbox}

\begin{promptbox}{Retrieval-Augmented Prompting baseline}{prompt:retrieval}
\promptpart{SYSTEM}
\begin{Verbatim}[breaklines=true]
You are a research forecasting assistant grounded in retrieved representative literature.
Return only valid JSON matching the requested schema.
\end{Verbatim}
\promptpart{USER}
\begin{Verbatim}[breaklines=true]
You are forecasting research ideas with a literature cutoff of @slot|cutoff_month^.
The following representative papers were retrieved from the historical literature as the
most relevant grounding for your forecast:
@slot|retrieved_block^
Based on these retrieved papers, forecast @slot|top_k^ concrete research ideas likely to
appear in the months following @slot|cutoff_month^. Do NOT use knowledge from after
@slot|cutoff_month^.
Return JSON only:
{"ideas": [{"title": "...", "rationale": "...", "approach": "...", "confidence": 0.0,
            "key_terms": []}]}
\end{Verbatim}
\end{promptbox}

\begin{promptbox}{Memory-Augmented Prompting baseline}{prompt:memory}
\promptpart{COMPRESS/SYSTEM}
\begin{Verbatim}[breaklines=true]
You are a research analyst. Output only a bullet-point list, no preamble.
\end{Verbatim}
\promptpart{COMPRESS/USER}
\begin{Verbatim}[breaklines=true]
The following papers were published before @slot|cutoff_month^.
@slot|paper_block^
Summarize the 8 most important recurring research themes from these papers as concise
bullet points (one line each). Focus on methodological trends and open problems, not
individual papers.
\end{Verbatim}
\promptpart{FORECAST/SYSTEM}
\begin{Verbatim}[breaklines=true]
You are a research forecasting assistant with memory of long-term trends.
Return only valid JSON matching the schema.
\end{Verbatim}
\promptpart{FORECAST/USER}
\begin{Verbatim}[breaklines=true]
Historical Memory (recurring themes from earlier literature):
@slot|memory^
Recent abstracts (up to @slot|cutoff_month^):
@slot|recent_block^
Based on the historical memory and recent literature, forecast @slot|top_k^ concrete research
ideas likely to appear in the months after @slot|cutoff_month^. Do NOT use knowledge from
after @slot|cutoff_month^.
Return JSON only:
{"ideas": [{"title": "...", "rationale": "...", "approach": "...", "confidence": 0.0,
            "key_terms": []}]}
\end{Verbatim}
\end{promptbox}

\needspace{6\baselineskip}
\subsection{Evaluation (Retrieve-then-Judge) Prompts}

\begin{promptboxbrk}{Judge system prompt}{prompt:judge-sys}
\promptpart{SYSTEM}
\begin{Verbatim}[breaklines=true]
You are an expert scientific reviewer. You are given a PREDICTED research direction -- a
forecast made before the paper was published -- and a PUBLISHED paper. Your task is to
judge whether the published paper is a realization of that prediction.
Score the prediction-paper pair on three dimensions using a 0-3 scale:
PROBLEM_MATCH -- Does the paper address the same core research problem and goal?
  3 = Same core problem and goal -- the prediction and paper tackle the exact same question
  2 = Very similar problem: same domain challenge, minor difference in scope or framing
  1 = Adjacent problem: overlapping concern but different primary objective
  0 = Unrelated or only superficially connected problem
METHOD_MATCH -- Does the paper employ a similar technical mechanism or approach?
  3 = Same or near-identical mechanism -- the prediction's approach is directly implemented
  2 = Closely related approach: shares key technical ideas, differs in implementation detail
  1 = Broadly similar paradigm: same general category of method, substantially different specifics
  0 = Different technical approach entirely
SPECIFICITY -- Does the paper realize the specific novelty described in the prediction?
  3 = Prediction's specific novelty is precisely present in the paper
  2 = Core specific novelty partially realized; paper simplifies or focuses on a subset
  1 = Prediction is generic enough to loosely fit, or paper addresses adjacent specifics
  0 = Prediction is keyword-only or meta-analytic, or paper entirely ignores the predicted novelty
A prediction MATCHES if (PROBLEM_MATCH + METHOD_MATCH >= 5) AND (SPECIFICITY >= 2).
Do NOT score based on shared topic or keyword overlap alone.
Here are four reference examples ordered from clear non-match to clear match:
--- EXAMPLE 1 (does NOT match -- paradigm overlap only) ---
Predicted Research Direction:
Title: Multiscale Generative Models for Long-Form Audio and Music with Hybrid
Token-Spectrogram Representations
Rationale: Diffusion models show promise for long high-fidelity music, and multiscale
autoregressive architectures have proven effective for very long sequences. A likely
near-term direction is combining hierarchical temporal structure with efficient
long-context generation specifically for raw or near-raw audio.
Approach: Use a two-level model where a global transformer predicts coarse musical
structure over long spans, and local modules generate fine-grained spectrogram patches
or audio tokens. Explore diffusion or autoregressive local decoders with recurrent
memory for minute-scale coherence.
Key Terms: music generation, audio modeling, multiscale transformer, diffusion,
hierarchical generation, long-range coherence
Published Paper:
Title: MEGABYTE: Predicting Million-byte Sequences with Multiscale Transformers
Abstract: We propose MEGABYTE, a multiscale decoder architecture for end-to-end modeling
of sequences over one million bytes. It segments sequences into patches, using a local
submodel within patches and a global model between patches. Experiments show competitive
performance on long-context language modeling, image density estimation, and audio from
raw files.
PROBLEM_MATCH: 1
METHOD_MATCH: 2
SPECIFICITY: 1
REASONING: The prediction targets audio and music generation as its primary objective,
while MEGABYTE is a general-purpose byte-sequence architecture whose audio experiments
are incidental -- different primary objectives. Both share a global-local hierarchy, so
the method is broadly similar, but the prediction's specific novelties (spectrogram
tokens, diffusion decoders, music-specific design) are entirely absent from MEGABYTE's
domain-agnostic patch approach.
--- EXAMPLE 2 (does NOT match -- mechanism diverges) ---
Prediction: "Hierarchical Segment-Level Memory Routing with Learned Boundary Detection
for Infinite-Context LLMs" ...
Paper: "SnapKV", which retains important KV entries observed over a fixed prefix window. ...
PROBLEM_MATCH: 2
METHOD_MATCH: 1
SPECIFICITY: 0
REASONING: Both address long-context efficiency via selective KV retention, but the
prediction's learned boundary detector, hierarchical routing, and cross-segment
attention are absent from SnapKV's fixed-window heuristic.
--- EXAMPLE 3 (MATCHES -- core idea realized, details differ) ---
Prediction: "Adaptive KV Cache Compression via Importance Prediction and Mixed
Precision". ...
Paper: "Scissorhands", which evicts non-pivotal tokens under a fixed memory budget. ...
PROBLEM_MATCH: 3
METHOD_MATCH: 2
SPECIFICITY: 2
REASONING: Both reduce KV cache memory via importance-based eviction, so the core idea
is present; the prediction's trained per-head importance predictor and mixed-precision
storage are not.
--- EXAMPLE 4 (MATCHES -- exact realization) ---
Prediction: "Distilling Explicit Chain-of-Thought into Implicit Latent Reasoning". ...
Paper: "Implicit Chain of Thought Reasoning via Knowledge Distillation". ...
PROBLEM_MATCH: 3
METHOD_MATCH: 3
SPECIFICITY: 3
REASONING: Both internalize chain-of-thought into latent states via teacher-student
distillation with no intermediate tokens at inference -- problem, mechanism, and goal
are identical.
---
Respond with exactly four lines:
PROBLEM_MATCH: <0, 1, 2, or 3>
METHOD_MATCH: <0, 1, 2, or 3>
SPECIFICITY: <0, 1, 2, or 3>
REASONING: <one to two sentences explaining your scores>
\end{Verbatim}
\end{promptboxbrk}

\begin{promptbox}{Judge user template}{prompt:judge-user}
\promptpart{USER}
\begin{Verbatim}[breaklines=true]
## Predicted Research Direction
Title: @slot|pred_title^
Rationale: @slot|pred_rationale^
Approach: @slot|pred_approach^
Key Terms: @slot|pred_terms^
## Published Paper
Title: @slot|paper_title^
Abstract: @slot|paper_abstract^
Score this prediction-paper pair on PROBLEM_MATCH, METHOD_MATCH, and SPECIFICITY (0-3
each), then give REASONING.
\end{Verbatim}
\end{promptbox}

\end{document}